%% file: neurips_2026.tex
\documentclass{article}
\PassOptionsToPackage{numbers}{natbib}
\usepackage[preprint]{neurips_2026}

\usepackage{bbding}
\usepackage[utf8]{inputenc}
\usepackage[T1]{fontenc}
\usepackage{url}
\usepackage{booktabs}
\usepackage{amsfonts}
\usepackage{amsmath}
\usepackage{amssymb}
\usepackage{amsthm}
\usepackage{nicefrac}
\usepackage{microtype}
\usepackage{graphicx}
\usepackage{subcaption}
\usepackage{wrapfig}
\usepackage{multirow}
\usepackage{algorithm}
\usepackage{algorithmic}
\usepackage{enumitem}
\usepackage{soul}
\usepackage{bbm}
\usepackage{bm}
\usepackage{makecell}
\usepackage{colortbl}
\usepackage{pifont}
\usepackage{tikz}
\usetikzlibrary{shapes,arrows,positioning,fit,backgrounds,shadows}
\usepackage{fontawesome5} 

\usepackage[dvipsnames]{xcolor}
\usepackage[table]{xcolor}
\theoremstyle{plain}

\theoremstyle{definition}

\theoremstyle{remark}

\definecolor{scrcolor}{HTML}{FF9100}   
\definecolor{ssrcolor}{HTML}{7B61FF}   
\definecolor{mcrcolor}{HTML}{00B8D4}   
\definecolor{deltacolor}{RGB}{130,81,223} 

\definecolor{scholarblue}{rgb}{0.21,0.49,0.74}
\definecolor{cornellred}{rgb}{0.7,0.11,0.11}
\definecolor{cadmiumgreen}{rgb}{0.0,0.42,0.24}
\definecolor{Blue9}{rgb}{0.098,0.3,0.9}
\definecolor{good}{RGB}{34,139,34}
\definecolor{bad}{RGB}{220,20,60}

\usepackage[
    colorlinks=true,
    linkcolor=red,
    citecolor=blue,
]{hyperref}

\usepackage[most]{tcolorbox}

\newcommand{\styledquote}[3][scholarblue!8]{%
\begin{center}
\colorbox{#1}{%
    \hspace{0.1in}%
    \begin{minipage}{0.85\textwidth}
    \vspace{0.1in}
    \small\itshape
    #2
    \def\temp{#3}%
    \ifx\temp\empty\else
        \begin{flushright}
        \small\normalfont ---#3
        \end{flushright}
    \fi
    \vspace{0.1in}
    \end{minipage}%
    \hspace{0.1in}%
}
\end{center}
}

\newcommand{\xmark}{\ding{55}}

\newcommand{\pae}{\textsc{PAE}}
\newcommand{\scr}{\textsc{SCR}}
\newcommand{\ssr}{\textsc{SSR}}
\newcommand{\mcr}{\textsc{MCR}}

\newcommand{\reals}{\mathbb{R}}

\newcommand{\loss}{\mathcal{L}}

\newcommand{\latent}{\mathbf{z}}

\newcommand{\fref}[1]{Fig.~\ref{#1}}

\title{What Matters for Diffusion-Friendly Latent Manifold? \\ Prior-Aligned Autoencoders for Latent Diffusion}

\author{%
  \textbf{Zhengrong Yue}$^{1,2}$, 
  \textbf{Taihang Hu}$^{2}$, 
  \textbf{Mengting Chen}$^{2,\dagger}$, 
  \textbf{Haiyu Zhang}$^{4}$, 
  \textbf{Zihao Pan}$^{5,2}$, \\
  \textbf{Tao Liu}$^{6,2}$, 
  \textbf{Zikang Wang}$^{1}$, 
  \textbf{Jinsong Lan}$^{2}$, 
  \textbf{Xiaoyong Zhu}$^{2}$, 
  \textbf{Bo Zheng}$^{2,\text{\Envelope}}$, 
  \textbf{Yali Wang}$^{3,7,\text{\Envelope}}$ \\
  $^1$Shanghai Jiao Tong University, $^2$Alibaba Group, \\
  $^3$ Shenzhen Institutes of Advanced Technology, Chinese Academy of Sciences, \\
  $^4$Beihang University, $^5$Sun Yat-sen University, $^6$Nankai University, $^7$Shanghai AI Laboratory \\
  \small $^\dagger$\ Project Leader, \Envelope\ Corresponding Author
}
\vspace{-0.5cm}

\begin{document}
\maketitle




\begin{center}
    \vspace{-2em} 
    \small 
    \href{https://github.com/ZhengrongYue/PAE}{%
        \includegraphics[height=1em]{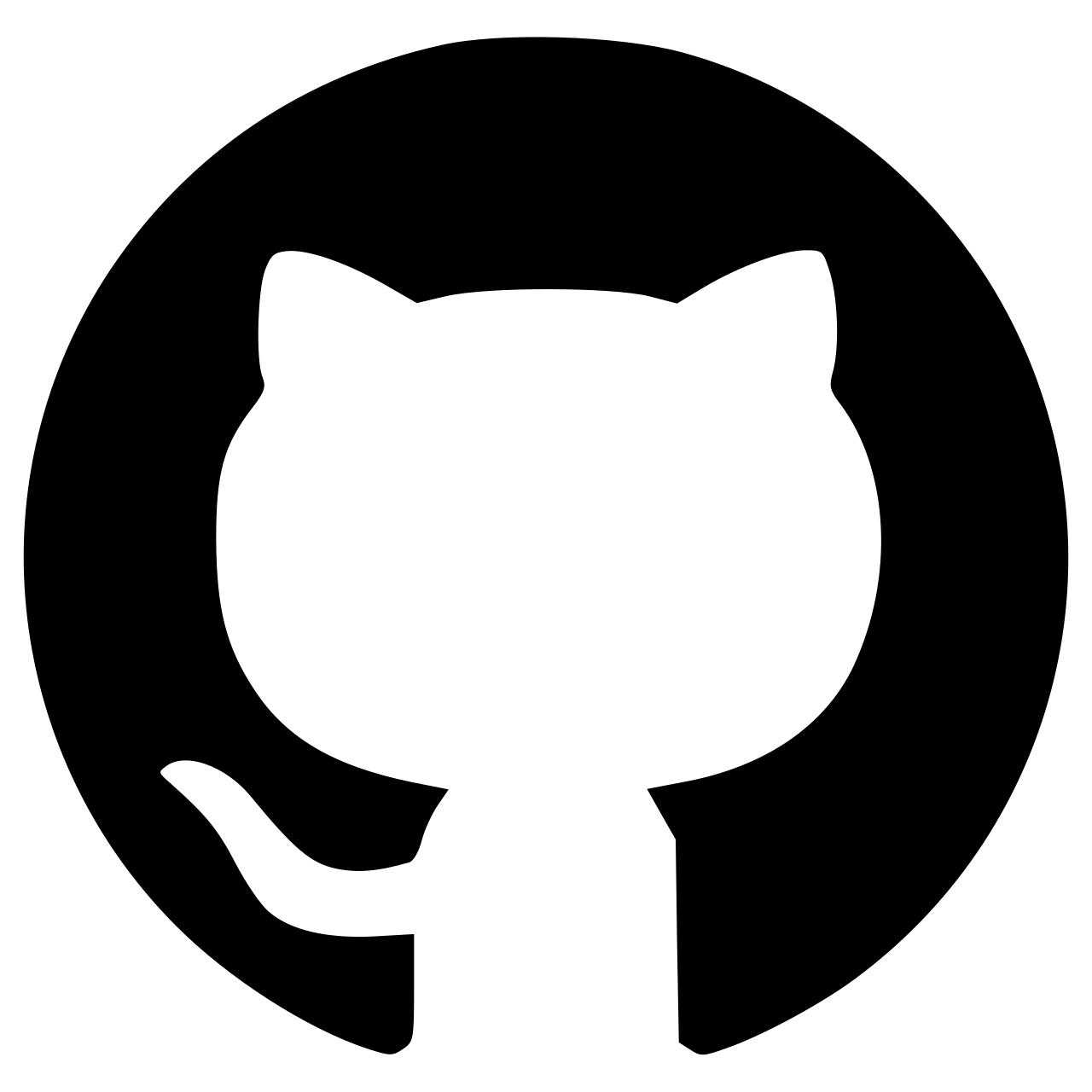}\,\texttt{Code}%
    }%
    \hspace{2em} 
    \href{https://huggingface.co/yuezhengrong/PAE-collections}{%
        \includegraphics[height=1em]{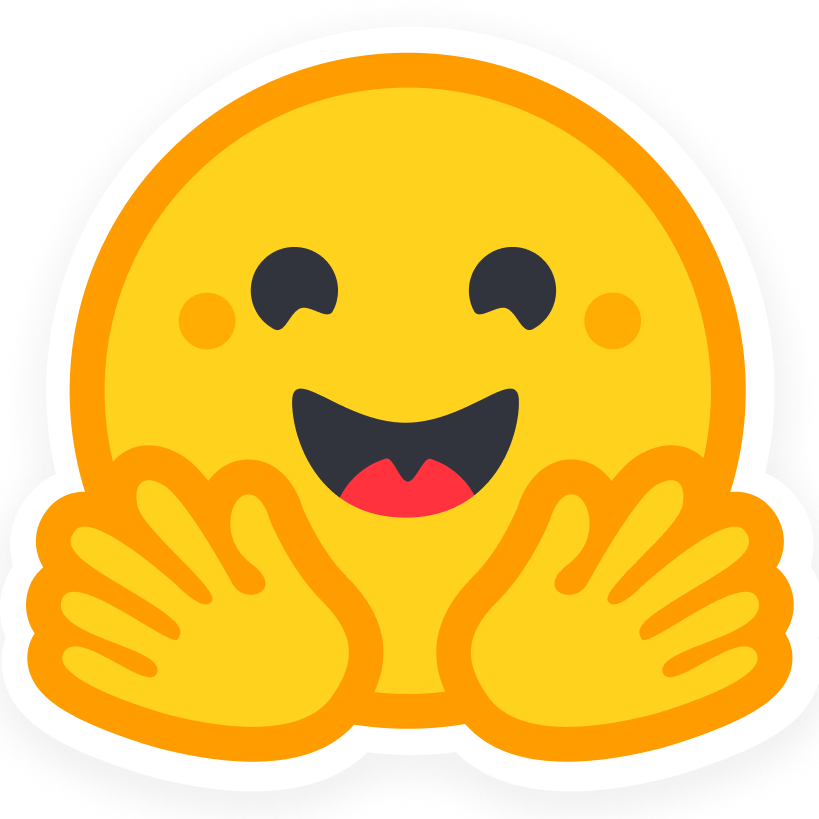}\,\texttt{HuggingFace}%
    }%
    \hspace{2em} 
    \href{https://www.modelscope.cn/models/ZhengrongYue/PAE-Collections}{%
        \includegraphics[height=0.9em]{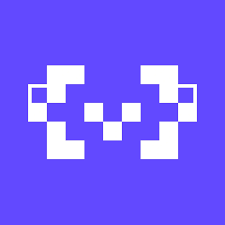}\,\texttt{ModelScope}%
    }
\end{center}
\begin{abstract}
Tokenizers are a crucial component of latent diffusion models, as they define the latent space in which diffusion models operate. However, existing tokenizers are primarily designed to improve reconstruction fidelity or inherit pretrained representations, leaving unclear what kind of latent space is truly friendly for generative modeling. In this paper, we study this question from the perspective of latent manifold organization. By constructing controlled tokenizer variants, we identify three key properties of a diffusion-friendly latent manifold: coherent spatial structure, local manifold continuity, and global manifold semantics. We find that these properties are more consistent with downstream generation quality than reconstruction fidelity. Motivated by this finding, we propose the \textit{\textbf{P}rior-\textbf{A}ligned Auto\textbf{E}ncoder} (\textbf{PAE}), which explicitly shapes the latent manifold instead of leaving diffusion-friendly manifold to emerge indirectly from reconstruction or inheritance. Specifically, PAE leverages refined priors derived from VFMs
and perturbation-based regularization to turn spatial structure, local continuity, and global semantics into explicit training objectives. 
On ImageNet $256{\times}256$, PAE improves both training efficiency and generation quality over existing tokenizers, reaching performance comparable to RAE with up to 13× faster convergence under the same training setup and achieving a new state-of-the-art gFID of \textbf{1.03}. These results highlight the importance of organizing the latent manifold for latent diffusion models.
\end{abstract}

\begin{center}
    \vspace{-0.4cm}
    \includegraphics[width=\textwidth]{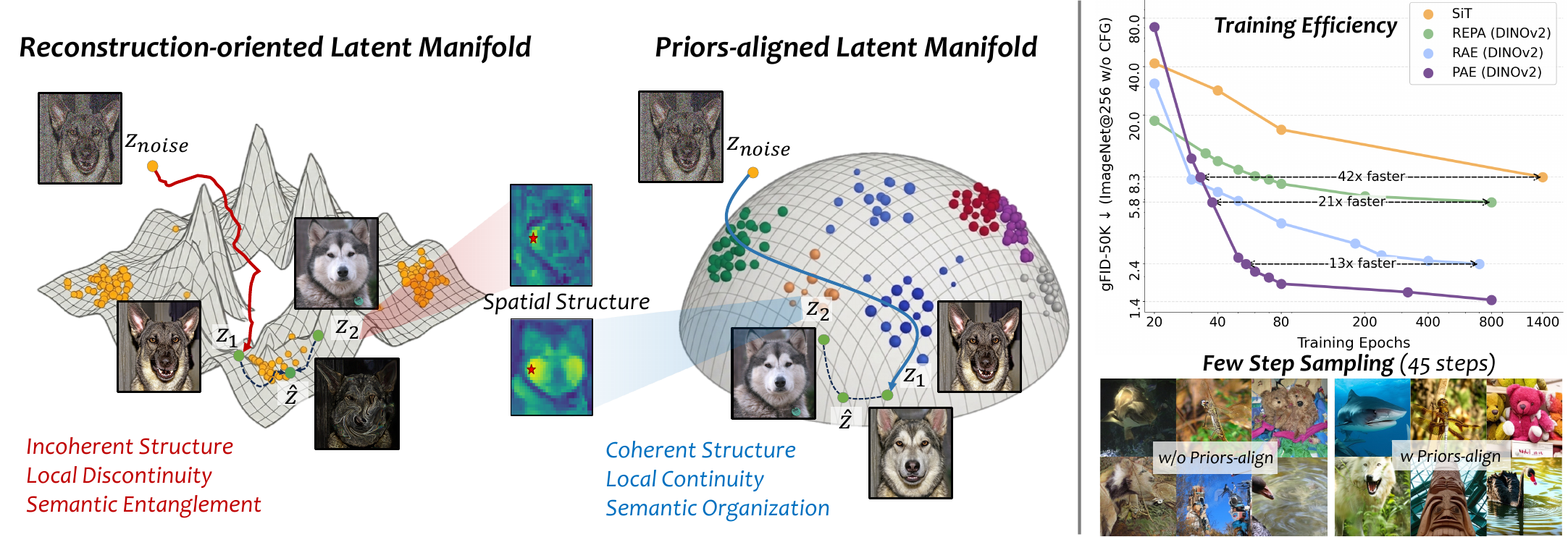}
    \captionof{figure}{
    \textbf{Prior alignment constructs a diffusion-friendly latent manifold.} 
    \textit{Left:} a conceptual illustration of latent space under the manifold assumption~\cite{manifold_with_ldm}. Compared with the reconstruction-oriented counterpart, the prior-aligned latent manifold is more structurally coherent, locally continuous, and semantically organized.
    \textit{Right:} \pae{} yields faster convergence, better generation quality, and robust few-step sampling performance.}
    \label{fig:teaser}
\end{center}

\input{sections/introduction}
\input{sections/related_work}

\input{sections/method}
\input{sections/experiments}

\input{sections/conclusion}

\bibliographystyle{plainnat}
\bibliography{neurips_2026}
\clearpage

\newpage
\appendix
\input{sections/appendix}
\clearpage


\end{document}

%% file: sections/introduction.tex
\section{Introduction}
\label{sec:intro}
\vspace{-0.1cm}


\begin{figure}[t]
    \centering
    \includegraphics[width=\textwidth]{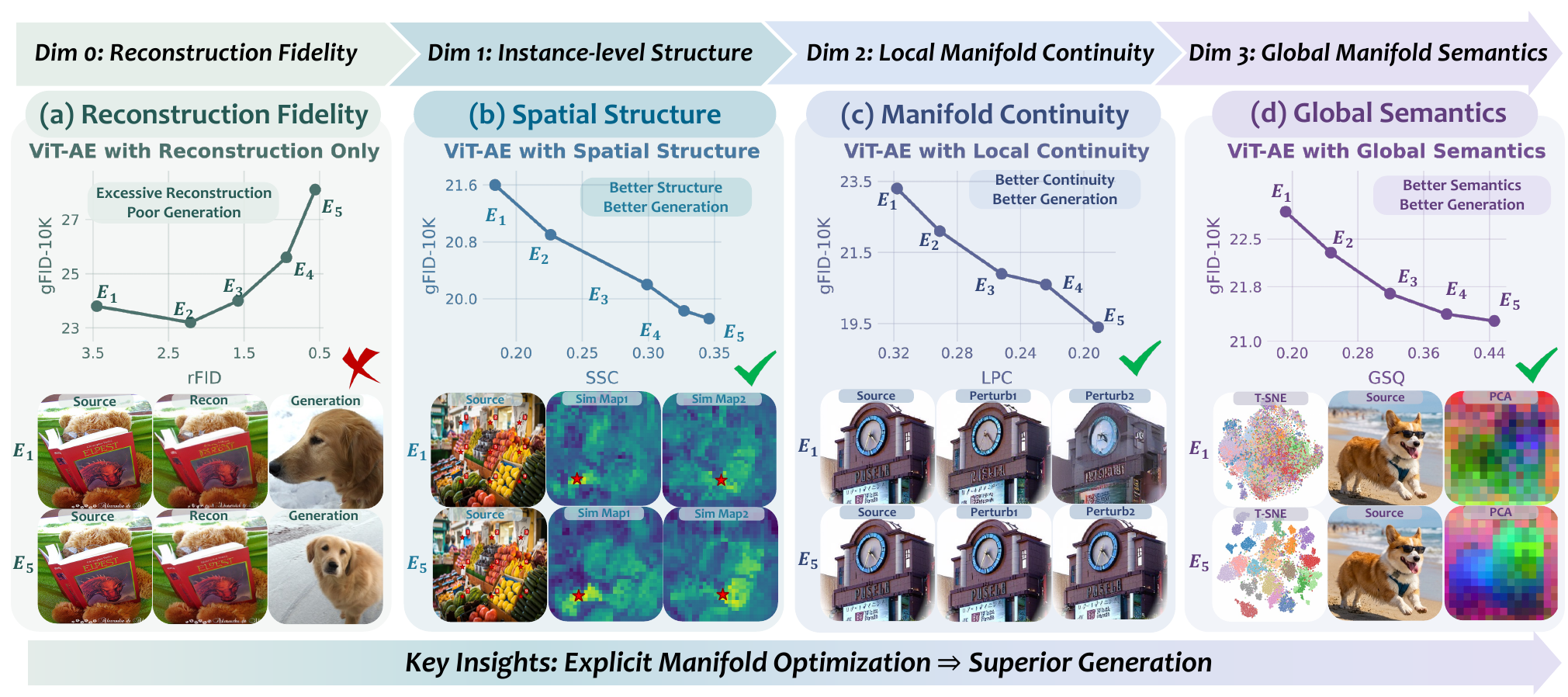}
    \vspace{-0.6cm}
    \caption{\textbf{Pilot experiments on diffusion-friendly latent manifold properties.}
    (a) Better reconstruction alone (rFID) does not guarantee better generation quality (gFID).
    (b--d) In contrast, improvements in instance-level structure, local manifold continuity, and global manifold semantics consistently correlate with better generation across controlled tokenizer variants. Together, these motivate latent-manifold organization as an explicit objective for designing tokenizers. Full settings and metric definitions are provided in Appendix~\ref{app:metrics_theory}.}  
    \vspace{-0.6cm}
    \label{fig:pilot}
\end{figure}



Latent diffusion models (LDMs)~\citep{ldm,sdxl,dit} achieve high-fidelity image synthesis by performing diffusion in a compressed latent space, substantially reducing computational cost while preserving visual detail.
As shown in \fref{fig:teaser}, the compressed latent space plays a crucial role in both the training efficiency and generation quality of diffusion models, underscoring the requirement for constructing a diffusion-friendly latent manifold~\citep{manifold_with_ldm}.


Vanilla variational autoencoder (VAE)~\citep{vae} is optimized with a pixel-wise reconstruction loss and the KL regularization term. While this reconstruction-oriented objective enables high-quality reconstruction, it can induce a \textit{reconstruction-generation mismatch}~\citep{vavae}. As illustrated in \fref{fig:pilot}(a), improving reconstruction performance alone does not necessarily lead to better generation quality.


Recent studies have begun to move beyond reconstruction-oriented objectives by incorporating more structured representation priors from Vision Foundation Models (VFMs). A line of work directly adopts pretrained VFM features as the latent representation for diffusion~\citep{rae,fae}. While such features effectively preserve semantic structure and thus simplify generative modeling, their highly semantic abstraction makes it difficult to generate high-frequency details and perform fine-grained editing. Another line of work leverages VFMs as teachers to supervise the training of tokenizers via feature alignment or distillation~\citep{gae,uniflow,aligntok}. While these methods can 
inherit useful semantic priors from teacher models and enhance the generation of high-frequency details, they provide limited analysis of how the latent space should be organized. This leaves a fundamental question: \textbf{\textit{what kind of latent space is actually friendly for diffusion?}}

To fill this gap, we analyze the problem from the perspective of \textit{latent manifold construction}~\citep{manifold_secrets_for_ldm,manifold_with_ldm}, 
which aims to construct a more effective latent manifold that facilitates diffusion model learning. 
We conduct controlled pilot experiments (\fref{fig:pilot}) to investigate three complementary manifold properties: \textbf{(i)~Spatial Structure Coherence (SSC)} measures the spatial structure of each latent in terms of intra-instance similarity and inter-instance discriminability.  Improving this property enables the diffusion model to focus on learning generative patterns rather than compensating for spatial misalignment (\fref{fig:pilot}(b)). \textbf{(ii)~Local Perceptual Continuity (LPC)} quantifies the local Lipschitz continuity of the latent manifold by evaluating perceptual changes among neighboring decoded samples along interpolation paths. A locally continuous manifold provides smoother prediction targets for the diffusion model, benefiting both training convergence and inference efficiency (\fref{fig:pilot}(c)). \textbf{(iii)~Global Semantic Quality (GSQ)} captures how compactly data with similar semantic concepts are organized on the latent manifold. By clustering semantically similar samples, it endows the diffusion model with a globally semantic latent manifold, making conditional generation easier to learn. 
Throughout these controlled studies, we fix the latent channel budget and use \textit{eRank} (Appendix ~\ref{app:metric_definitions}) only as a supplementary diagnostic of latent utilization, so that the observed trends are mainly attributed to differences in manifold geometry. Our experiments show that these three manifold properties are strongly correlated with downstream gFID, suggesting that they serve as effective indicators of a diffusion-friendly latent manifold.

Inspired by these findings, we propose the \textit{\textbf{P}rior-\textbf{A}ligned Auto\textbf{E}ncoder} (\textbf{PAE}), a tokenizer that explicitly shapes the latent manifold. Specifically, we propose three targeted regularizations corresponding to the three manifold properties above: Spatial Structure Regularization (SSR) enhances instance-level spatial structure by aligning each latent with its corresponding VFM feature; Manifold Continuity Regularization (MCR) promotes local manifold continuity by perturbing latents and enforcing perceptual consistency between the decoded outputs; and Semantic Consistency Regularization (SCR) preserves global manifold semantics by aligning the latent manifold with globally pooled VFM features. However, VFM features can be channel-redundant for semantic supervision and spatially imprecise at the tokenizer resolution. Therefore, we introduce a lightweight projector that maps VFM features into the tokenizer resolution. We further upsample the VFM features and apply low-pass spatial refinement to obtain fine-grained alignment targets. In addition, the encoder of our tokenizer integrates a frozen VFM and a Detail-aware Modulator (DAM), improving training efficiency while enhancing the model’s capacity for modeling high-frequency details.

Experiments on ImageNet (256$\times$256) demonstrate that PAE improves both tokenizer quality and downstream diffusion generation. Our tokenizer achieves strong reconstruction performance with an rFID of 0.26. Under the same LightningDiT setting, PAE reaches performance comparable to RAE with up to $13\times$ fewer training epochs, as shown in Fig.~\ref{fig:teaser}. With longer training, it further establishes a new state-of-the-art gFID of 1.03. Moreover, PAE maintains generation quality with only 45 denoising steps, achieving a gFID of 1.05. More broadly, our results suggest a simple principle for tokenizer design: \textit{latent diffusion benefits from better diffusion-friendly manifold organization}.

\vspace{-0.3cm}

%% file: sections/related_work.tex
\section{Related Work}
\label{sec:related}
\vspace{-0.1cm}

\textbf{Representation Priors in Diffusion Generators.}
This paradigm is referred to as \textit{Representation-Guided DiT}, as it improves diffusion by injecting external representation priors into the generator.
Recent work improves diffusion training by reshaping generator-side representations. One line aligns DiT features with vision foundation model (VFM) representations~\cite{repa,repa-e,irepa}; another modifies the denoising process to model high-level semantics before pixel-level synthesis~\cite{reg,sfd,redi,latent_forcing}. Despite their differences, both directions operate on a fixed autoencoder-induced latent space. They improve how the generator models a given reconstruction-oriented representation space, rather than how that space should be constructed.

\textbf{Representation Autoencoders for Latent Diffusion.}
This paradigm is referred to as \textit{Representation-Native DiT}, as it improves downstream diffusion by constructing a representation-rich latent space through the autoencoder.
Latent diffusion relies on a first-stage autoencoder to define the latent space for downstream diffusion~\cite{ldm,vae}. Early VAE-based designs mainly optimize reconstruction fidelity~\cite{vae,ldm,sdxl,flux,dcae,sana}, but reconstruction quality alone is an insufficient proxy for generative performance~\cite{vavae}. This has motivated autoencoders with stronger representation priors, either by reconstructing frozen VFM features~\cite{rae,fae,svg,vfmvae,repack} or by distilling pretrained representations through alignment or joint objectives~\cite{gae,uniflow,psvae,aligntok,maetok,vtp,unitok}. While these methods enrich latent representations with pretrained structure, they mainly focus on inheriting or distilling stronger features. In contrast, PAE treats latent manifold construction itself as the primary objective of autoencoder design, rather than feature inheritance.
\vspace{-0.3cm}

%% file: sections/method.tex
\section{Method}
\label{sec:method}
\vspace{-0.1cm}
We propose \pae{}, a tokenizer framework improving latent diffusion by explicitly shaping the latent manifold beyond simple reconstruction. 
Using a frozen vision foundation model (VFM) as a semantic reference, \pae{} learns a compact space regularized along three diffusion-relevant dimensions: spatial structure, local continuity, and global semantics. 
Section~\ref{sec:arch} introduces the tokenizer architecture, followed by the prior alignment regularizations in Section~\ref{sec:objectives}. In Section~\ref{sec:refined_priors}, we introduce a refinement strategy for VFM features, enabling them to serve as more effective alignment targets for our regularizations.
\vspace{-0.1cm}

\subsection{PAE Architecture}
\label{sec:arch}
\vspace{-0.1cm}

\begin{figure}[t]
    \centering
    \includegraphics[width=\textwidth]{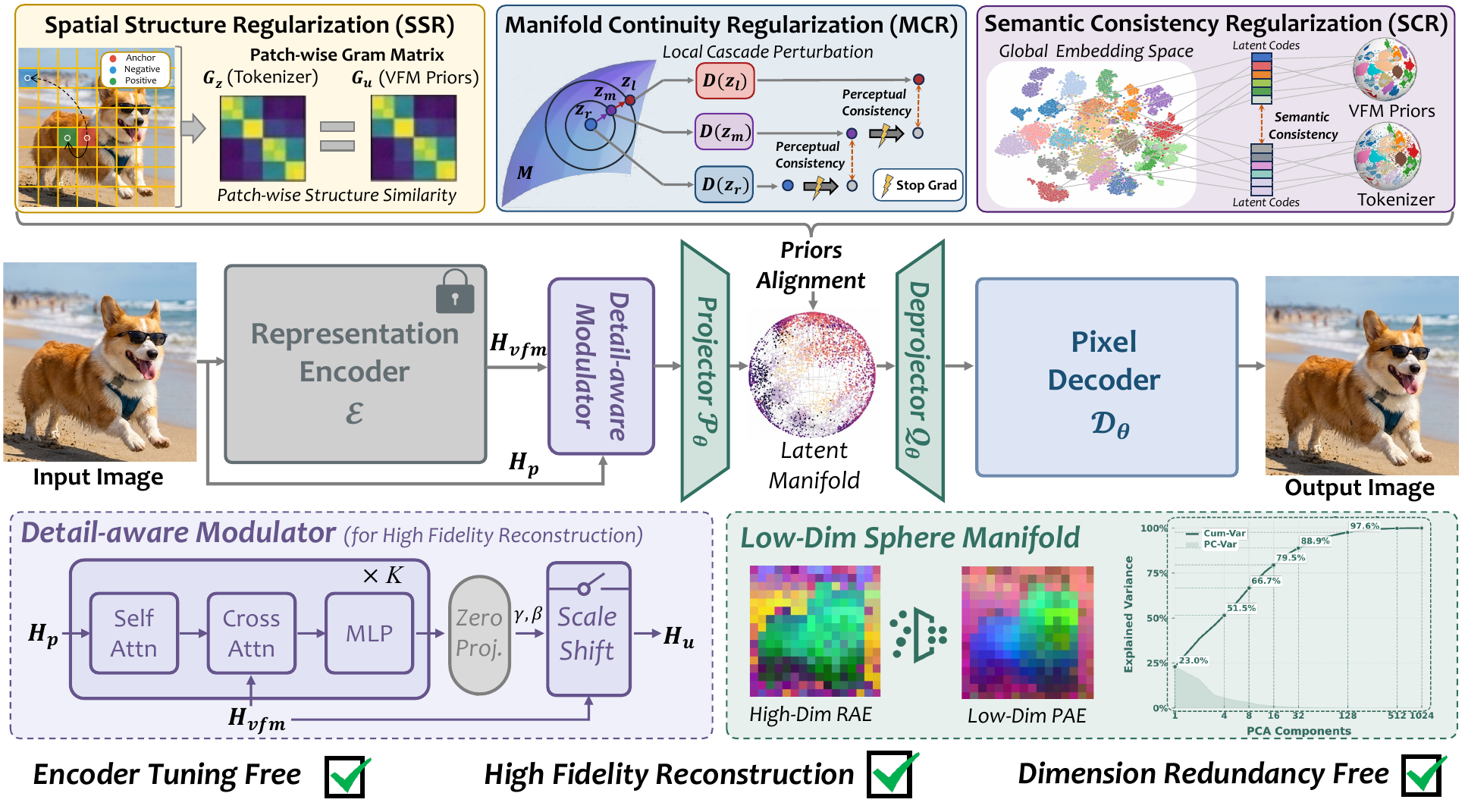}
    \caption{\textbf{Overview of the \pae{} framework.} A frozen VFM provides stable representation features for the input image. DAM injects pixel detail while preserving the VFM as the dominant semantic source. The modulated representation is projected into a compact latent space for downstream diffusion. On top of this backbone, three prior-alignment objectives explicitly shape the latent manifold: SSR preserves instance-level spatial structure, MCR enforces local continuity, and SCR preserves global semantic organization.}
    \vspace{-0.4cm}
    \label{fig:framework}
\end{figure}

\textbf{Overview.}
Given an input image $x \in \reals^{B \times 3 \times H \times W}$, \pae{} first extracts frozen VFM features $\mathbf{H}_{\mathrm{vfm}}=\mathcal{E}(x)\in\reals^{B \times N \times D}$. A lightweight modulator $\mathcal{DAM}_{\theta}(\mathbf{H}_{\mathrm{vfm}},x)$ then injects reconstruction-critical pixel detail into these frozen features. The modulated representation is projected into a compact latent code
$
z=\mathcal{P}_{\theta}(\mathcal{DAM}_{\theta}(\mathbf{H}_{\mathrm{vfm}},x))\in\reals^{B \times d \times H' \times W'},
$
which serves as the tokenizer output for downstream diffusion. 
For reconstruction, a deprojector $\mathcal{Q}_\theta$ maps $z$ back to representation space and a pixel decoder $\mathcal{D}_\theta$ reconstructs the image
$
\hat{x}=\mathcal{D}_{\theta}(\mathcal{Q}_{\theta}(z))\in\reals^{B \times 3 \times H \times W}.
$
Here $\mathcal{E}$ is frozen, while $\mathcal{DAM}_{\theta}$, $\mathcal{P}_{\theta}$, $\mathcal{Q}_{\theta}$, and $\mathcal{D}_{\theta}$ are trainable.

\textbf{Detail-Aware Modulator (DAM).}
Frozen VFM features provide a strong starting point but miss fine-grained visual detail needed for faithful reconstruction. Directly finetuning the VFM often weakens its pretrained structure. DAM addresses this by injecting pixel-level detail while keeping the frozen VFM features dominant. Specifically, we patchify the input image into pixel tokens $\mathbf{H}_p$ and process them through $K$ Transformer blocks as
$
\mathbf{H}_p^{(l)}
=
\text{MLP}\Big(
\text{CrossAttn}\big(
\text{SelfAttn}(\mathbf{H}_p^{(l-1)}),\mathbf{H}_{\mathrm{vfm}}
\big)\Big).
$
The output $\Delta \mathbf{H}=\mathbf{H}_p^{(K)}$ modulates the VFM features through zero-initialized scale-and-shift fusion,
\begin{equation}
    \boldsymbol{\gamma}_p, \boldsymbol{\beta}_p
    =
    \text{split}\big(\mathbf{W}\Delta \mathbf{H}\big), \qquad
    \mathbf{H}_{z}
    =
    \text{LayerNorm}\big(\mathbf{H}_{\mathrm{vfm}} \odot (1 + \boldsymbol{\gamma}_p) + \boldsymbol{\beta}_p\big),
    \label{eq:sft_new}
\end{equation}
where $\mathbf{W}$ is initialized to zero so that training starts from $\mathbf{H}_{z}=\mathbf{H}_{\mathrm{vfm}}$. This design gradually injects missing detail while preserving the pretrained VFM as the main semantic source, and avoids the uncontrolled mixing introduced by simple residual concatenation as~\cite{svg}.

\textbf{Low-dimensional Sphere Manifold.}
To derive a compact latent representation for downstream diffusion, the modulated representation $\mathbf{H}_{z}$ is projected as $\tilde{z}=\mathcal{P}_\theta(\mathbf{H}_z)$. Following the best practices in~\cite{gae}, the projector $\mathcal{P}_{\theta}$ consists of attention and convolution layers. To ensure a structured and navigable manifold, we normalize the compressed features by their root-mean-square (RMS) magnitude as $z=\tilde{z}/\sqrt{\mathrm{mean}(\tilde{z}^{\,2})+\epsilon} \in \reals^{B \times d \times H' \times W'}$,
where $z \in \mathbb{R}^{B \times d \times H' \times W'}$. This compact, sphere-like latent space not only enhances diffusion efficiency by removing channel redundancy but also stabilizes the local perturbations required for manifold continuity regularization.

\textbf{Decoding and Reconstruction.}
The deprojector $\mathcal{Q}_{\theta}$ maps the latent code $z$ back to representation space, after which the pixel decoder $\mathcal{D}_{\theta}$ reconstructs the image. Reconstruction is trained with
\begin{equation}
    \loss_{\text{recon}}
    =
    \loss_{\ell_1}
    + \lambda_{\text{lpips}}\loss_{\text{LPIPS}}
    + \lambda_{\text{gan}}\loss_{\text{GAN}}.
    \label{eq:recon}
\end{equation}
This ensures visual fidelity, but reconstruction alone does not produce a diffusion-friendly latent space. We therefore introduce prior alignment objectives to shape the latent manifold.

\vspace{-0.1cm}
\subsection{Prior Alignment Regularizations}
\label{sec:objectives}
\vspace{-0.1cm}
The core of \pae{} is to turn the three diffusion-friendly latent properties identified in our analysis into explicit training objectives. Beyond reconstruction, we regularize the latent space along three complementary dimensions: instance-level spatial structure, local continuity, and global semantic organization. For clarity, $\mathbf{Z}_T$ denotes the refined target feature from the frozen VFM in Sec.~\ref{sec:refined_priors}.

\textbf{Spatial Structure Regularization (SSR).} 
While strong reconstruction is essential, it does not guarantee that spatial relationships between latent tokens survive bottleneck compression. To preserve this instance-level topology, SSR aligns the spatial Gram matrices $\mathbf{G}_z = \mathbf{Z}^{\top}\mathbf{Z}$ and $\mathbf{G}_T = \mathbf{Z}_T^{\top}\mathbf{Z}_T$:
\begin{equation}
    \mathcal{L}_{\text{\ssr}} = \|\mathbf{G}_z - \mathbf{G}_T\|_F^2.
    \label{eq:ssr_new}
\end{equation}
This objective remains consistent with the relative structure prior for latent manifold.

\textbf{Manifold Continuity Regularization (MCR).}
Autoencoders mainly constrain reconstruction at observed data points, placing only weak pressure on nearby latent neighborhoods. A naive way to improve local robustness is to train the decoder to reconstruct from perturbed latents directly, but this typically introduces a trade-off: large perturbations can harm reconstruction fidelity, while very small perturbations provide only weak continuity regularization. MCR instead regularizes local smoothness most relevant to downstream diffusion through a cascaded perturbation consistency objective in latent space.
For each sample, let $\latent_r \sim q(\latent \mid x)$ be the reconstruction latent. We sample a direction $\Delta$ and construct two perturbed latents
\[
\latent_m = \latent_r + \alpha_m \Delta, \qquad
\latent_l = \latent_r + \alpha_l \Delta,
\qquad \alpha_l > \alpha_m > 0.
\]
For simplicity, we use $D(\cdot)$ to denote the full latent-to-image decoder, including the deprojector and the pixel decoder. Their reconstructions are $\hat{x}_r = D(\latent_r)$, $\hat{x}_m = D(\latent_m)$, and $\hat{x}_l = D(\latent_l)$. Rather than forcing all perturbed latents to reconstruct the original image directly, MCR imposes consistency only between neighboring perturbation levels:
\begin{equation}
    \loss_{\mcr}
    =
    \underbrace{
    \|\hat{x}_m - \text{sg}(\hat{x}_r)\|_1
    + \text{LPIPS}(\hat{x}_m, \text{sg}(\hat{x}_r))
    }_{\text{medium}\rightarrow\text{recon}}
    +
    \underbrace{
    \|\hat{x}_l - \text{sg}(\hat{x}_m)\|_1
    + \text{LPIPS}(\hat{x}_l, \text{sg}(\hat{x}_m))
    }_{\text{large}\rightarrow\text{medium}}.
    \label{eq:mcr_new}
\end{equation}
Here $\text{sg}(\cdot)$ denotes stop-gradient. This cascaded design regularizes the local latent neighborhood in a progressive and less destructive manner, encouraging nearby latent points to decode to perceptually similar images while preserving the reconstruction quality of the anchor latent.

\begin{figure*}[t]
\vspace{-0.6cm}
    \centering
    \includegraphics[width=\textwidth]{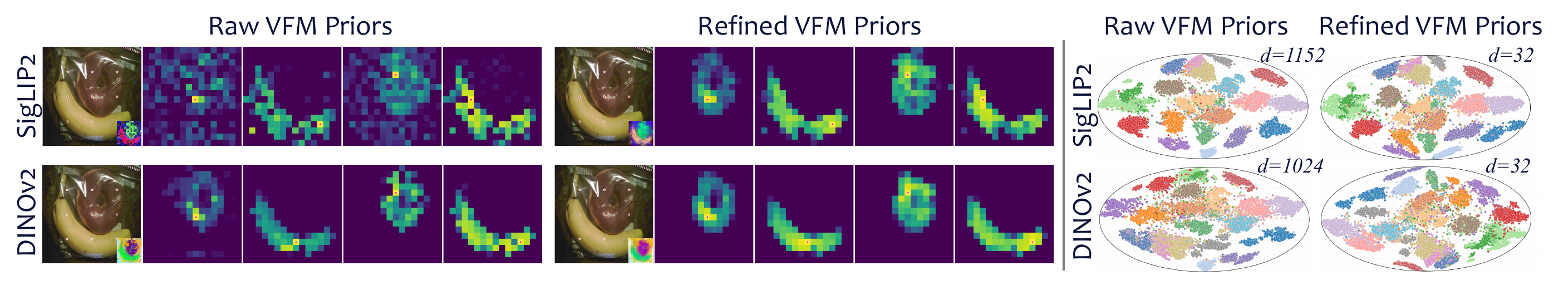}
    \vspace{-0.4cm}
    \caption{\textbf{Refined VFM priors provide better-matched alignment targets for \pae{}.}
    \textit{Left}: refined structural targets exhibit clearer patch-wise spatial correlations, yielding cleaner supervision for SSR.
    \textit{Right}: compressed semantic targets remain well clustered in embedding space, indicating improved bottleneck matching without losing semantic organization.}
    \label{fig:refine_vfm}
    \vspace{-0.4cm}
\end{figure*}

\textbf{Semantic Consistency Regularization (SCR).}
Bottleneck compression can distort the semantic directions inherited from pretrained representations. SCR preserves global semantic organization by aligning the compressed low-dimensional tokenizer tokens with the projected target tokens at both pooled and patch-token levels. Let $\mathbf{Z}_T$ denote the patch-level target tokens, $\mathbf{z}_{T,g}$ their pooled token, $\mathbf{Z}$ the compressed low-dimensional tokenizer tokens, and $\mathbf{z}_g$ the pooled token. The loss is
\begin{equation}
    \loss_{\scr}
    =
    \Big(
    1 - \cos(\bar{\mathbf{z}}_{T,g}, \bar{\mathbf{z}}_g)
    \Big)
    +
    \Big(
    1 - \cos(\bar{\mathbf{Z}}_T, \bar{\mathbf{Z}})
    \Big),
    \label{eq:scr_new}
\end{equation}
where $\bar{\cdot}$ denotes $\ell_2$ normalization. The first term preserves concept-level organization through pooled semantic alignment, while the second term preserves token-wise semantic directions in the compressed low-dimensional token space.

\textbf{Overall objective.}
The total prior alignment regularization is defined as
\begin{equation}
    \loss_{p}
    =
    \lambda_{ssr} \loss_{\ssr}
    + \lambda_{mcr} \loss_{\mcr}
    + \lambda_{scr} \loss_{\scr}.
    \label{eq:lp}
\end{equation}
The final training objective is $\loss_{total} = \loss_{\text{recon}} + \loss_{p}$.

\subsection{Refining VFM Priors}
\label{sec:refined_priors}
\vspace{-0.1cm}

The objectives above rely on fixed target features derived from the frozen VFM. However, raw VFM features are not directly suitable as alignment targets: they are channel-redundant as semantic supervision and spatially imperfect at tokenizer resolution. In particular, as also observed in~\cite{gae}, directly distilling high-dimensional VFM features into a compact latent bottleneck is often mismatched for semantic supervision. \textit{A useful VFM-derived target should remain semantically informative under a compact tokenizer bottleneck while providing cleaner spatial structure at tokenizer resolution.} We therefore refine the frozen VFM into bottleneck-matched targets before tokenizer training.

Concretely, we first learn a lightweight prior projector \(\mathcal{P}_{\theta}^{t}\) that compresses raw VFM features into a compact target feature \(\mathbf{Z}_T = \mathcal{P}_{\theta}^{t}(\mathbf{H}_{\mathrm{vfm}})\) while reconstructing the original high-dimensional representation, yielding a semantic target whose pooled summary \(\mathbf{z}_{T,g}\) preserves semantics but better matches the tokenizer bottleneck. In parallel, we refine the VFM feature spatially by upsampling it, applying low-pass spatial refinement, and downsampling it back to latent resolution, which suppresses noisy local variation while preserving coarse spatial relations for SSR. Both targets are fixed during tokenizer training. As shown in Fig.~\ref{fig:refine_vfm}, the refined structural target yields clearer patch-wise spatial correlations for structure alignment, while the compressed semantic target remains well organized in embedding space despite the reduced dimensionality, indicating improved bottleneck matching without losing class-level semantics. More implementation details are given in Appendix~\ref{app:stage0}.

\begin{figure*}[t]
    \centering
    \includegraphics[width=\textwidth]{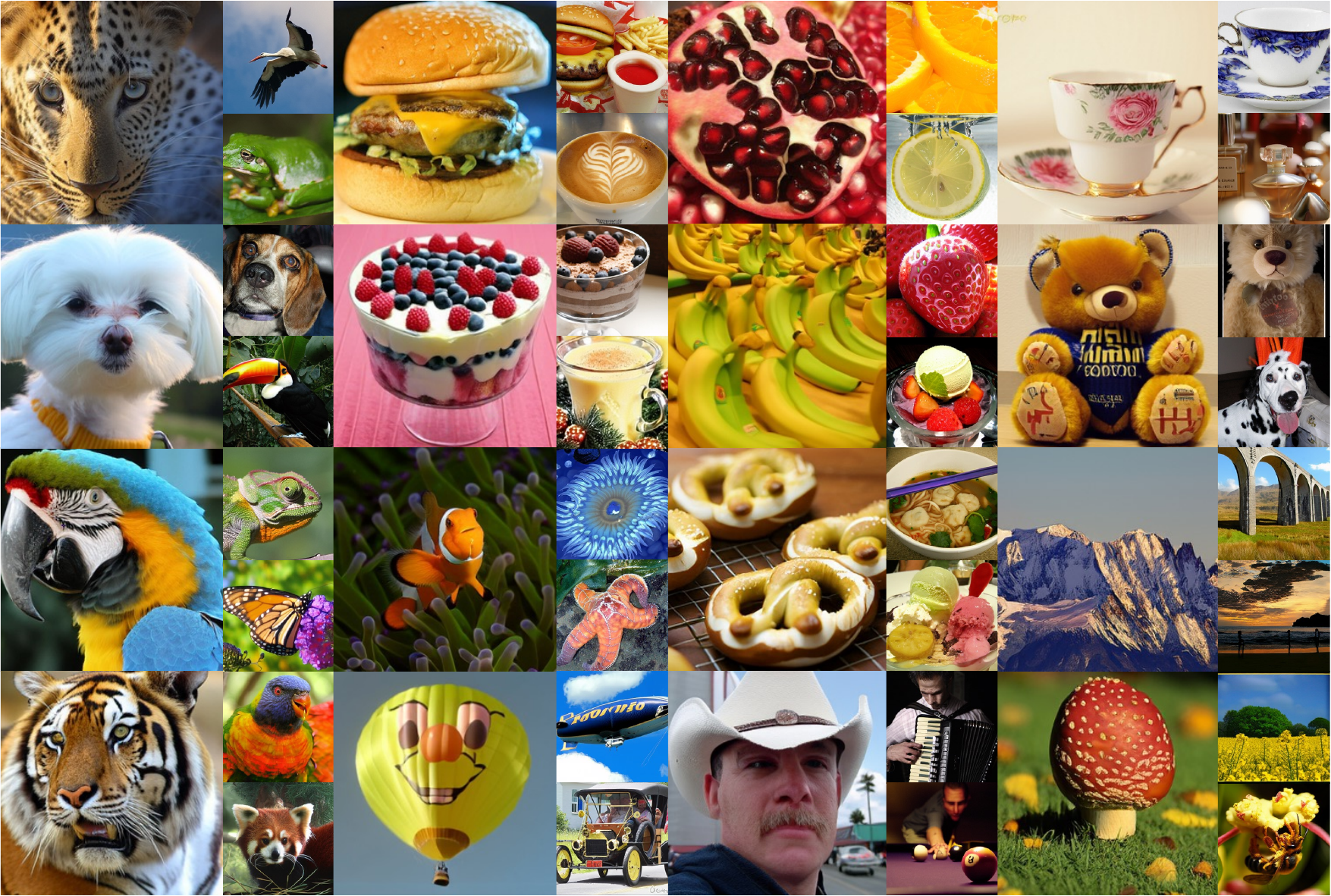}
    \vspace{-0.4cm}
    \caption{Class-conditional samples by PAE with LightningDiT-XL/1 show excellent image quality.}
    \label{fig:gen_vis}
    \vspace{-0.4cm}
\end{figure*}


\vspace{-0.3cm}

%% file: sections/experiments.tex
\section{Experiments}
\label{sec:experiments}
\vspace{-0.2cm}


\begin{table*}[!t]
    \centering
    \small
    \caption{\textbf{Generation performance on ImageNet $256{\times}256$.} PAE improves both convergence efficiency and final generation quality under the same training setup. In particular, PAE (DINOv2) achieves 1.27 gFID at 80 epochs and a new state-of-the-art 1.03 gFID at 800 epochs. $^*$ indicates results obtained with AutoGuidance~\cite{autoguidance} as reported in the original work.}
    \label{tab:main_system}
    \renewcommand{\arraystretch}{1.1}
    \setlength{\tabcolsep}{4pt}
    \resizebox{\textwidth}{!}{%
    \begin{tabular}{l c c c cccc cccc}
    \toprule
    \multirow{2}{*}{\textbf{Method}} &
    \textbf{Tokenizer} &
    \multirow{2}{*}{\textbf{\makecell{Generator\\Params}}} &
    \multirow{2}{*}{\textbf{Epochs}} &
    \multicolumn{4}{c}{\textbf{Generation@256 w/ Guidance}} &
    \multicolumn{4}{c}{\textbf{Generation@256 w/o Guidance}} \\
    \cmidrule(lr){5-8} \cmidrule(lr){9-12}
    & \textbf{rFID}$\downarrow$
    & &
    &
    \textbf{gFID}$\downarrow$
    & \textbf{IS}$\uparrow$
    & \textbf{Prec.}$\uparrow$
    & \textbf{Rec.}$\uparrow$
    & \textbf{gFID}$\downarrow$
    & \textbf{IS}$\uparrow$
    & \textbf{Prec.}$\uparrow$
    & \textbf{Rec.}$\uparrow$ \\
    \midrule

    \multicolumn{12}{c}{\textit{\textbf{Convergence Efficiency for Representation-Guided DiT}}} \\
    \midrule
    DDT~\cite{ddt} & 0.61 & 675M & 80 & 1.52 & 263.7 & 0.78 & 0.63 & 6.62 & 135.2 & 0.69 & 0.67 \\
    REPA~\cite{repa} & 0.61 & 675M & 80 & 1.42 & 305.7 & 0.80 & 0.65 & 7.90 & 122.6 & 0.70 & 0.65 \\
    REPA-E~\cite{repa-e} & 0.28 & 675M & 80 & 1.67 & 266.3 & 0.80 & 0.63 & 3.46 & 159.8 & 0.77 & 0.63 \\
    REG~\cite{reg} & 0.58 & 675M & 80 & 1.86 & 321.4 & 0.76 & 0.66 & 3.40 & 184.1 & -- & -- \\
    \textcolor{gray}{SFD$^*$~\cite{sfd}} & \textcolor{gray}{0.26} & \textcolor{gray}{675M} & \textcolor{gray}{80} & \textcolor{gray}{1.30} & \textcolor{gray}{233.4} & \textcolor{gray}{0.78} & \textcolor{gray}{0.65} & \textcolor{gray}{3.53} & \textcolor{gray}{--} & \textcolor{gray}{--} & \textcolor{gray}{--} \\
    \midrule

    \multicolumn{12}{c}{\textit{\textbf{Convergence Efficiency for Representation-Native DiT}}} \\
    \midrule
    SVG~\cite{svg} & 0.65 & 675M & 80 & 3.54 & 207.6 & -- & -- & 6.57 & 137.9 & -- & -- \\
    VA-VAE~\cite{vavae} & 0.28 & 675M & 64 & 2.11 & 252.3 & 0.81 & 0.58 & 5.14 & 130.2 & 0.76 & 0.62 \\
    VFM-VAE~\cite{vfmvae} & 0.52 & 675M & 80 & 2.16 & 232.8 & 0.82 & 0.58 & 3.80 & 152.8 & -- & -- \\
    AlignTok~\cite{aligntok} & 0.26 & 675M & 64 & 1.90 & 260.9 & 0.81 & 0.61 & 3.71 & 148.9 & 0.77 & 0.62 \\
    \textcolor{gray}{RAE (DiTDH-XL)$^*$~\cite{rae}} & \textcolor{gray}{0.57} & \textcolor{gray}{839M} & \textcolor{gray}{80} & \textcolor{gray}{--} & \textcolor{gray}{--} & \textcolor{gray}{--} & \textcolor{gray}{--} & \textcolor{gray}{2.16} & \textcolor{gray}{214.8} & \textcolor{gray}{0.82} & \textcolor{gray}{0.59} \\
    Send-VAE (w. REPA)~\cite{sendvae} & 0.31 & 675M & 80 & 1.41 & 301.7 & 0.79 & \textbf{0.65} & 2.88 & 175.3 & 0.78 & 0.62 \\
    RPiAE~\cite{rpiae} & 0.50 & 675M & 80 & 1.51 & 225.9 & 0.79 & \textbf{0.65} & 2.25 & 208.7 & 0.81 & 0.60 \\
    FAE~\cite{fae} & 0.68 & 675M & 80 & 1.70 & 243.8 & 0.82 & 0.61 & 2.08 & 207.6 & \textbf{0.82} & 0.59 \\
    GAE~\cite{gae} & 0.44 & 675M & 80 & 1.48 & 265.2 & 0.80 & 0.62 & 1.82 & \textbf{220.4} & \textbf{0.82} & 0.61 \\
    VTP~\cite{vtp} & 0.36 & 675M & 80 & 1.44 & 238.2 & 0.80 & 0.63 & 2.62 & 197.8 & 0.79 & 0.62 \\
    \rowcolor{blue!8}
    \textbf{PAE (MAE)} & \textbf{0.23} & 675M & 80 & 2.81 & \textbf{316.0} & \textbf{0.85} & 0.57 & 3.65 & 156.9 & 0.78 & 0.61 \\
    \rowcolor{blue!8}
    \textbf{PAE (SigLIP2)} & 0.27 & 675M & 80 & 1.39 & 268.3 & 0.79 & \textbf{0.65} & 2.32 & 199.6 & 0.81 & 0.62 \\
    \rowcolor{blue!8}
    \textbf{PAE (DINOv3)} & 0.28 & 675M & 80 & 1.31 & 262.7 & 0.78 & \textbf{0.65} & 1.81 & 216.7 & 0.80 & 0.62 \\
    \rowcolor{blue!8}
    \textbf{PAE (DINOv2)} & 0.26 & 675M & 80 & \textbf{1.27} & 275.3 & 0.79 & \textbf{0.65} & \textbf{1.80} & 218.3 & \textbf{0.82} & 0.62 \\
    \midrule

    \multicolumn{12}{c}{\textit{\textbf{Long Period Training for Representation-Guided DiT}}} \\
    \midrule
    DDT~\cite{ddt} & 0.61 & 675M & 400 & 1.26 & 310.6 & 0.79 & 0.65 & 6.27 & 154.7 & 0.68 & 0.69 \\
    REPA~\cite{repa} & 0.61 & 675M & 800 & 1.29 & 306.3 & 0.79 & 0.64 & 5.78 & 158.3 & 0.70 & 0.68 \\
    REPA-E~\cite{repa-e} & 0.28 & 675M & 800 & 1.15 & 304.0 & 0.79 & 0.66 & 1.70 & 217.3 & 0.77 & 0.66 \\
    ReDi~\cite{redi} & 0.58 & 675M & 800 & 1.61 & 295.1 & 0.78 & 0.64 & -- & -- & -- & -- \\
    REG~\cite{reg} & 0.58 & 675M & 800 & 1.36 & 299.4 & 0.77 & 0.66 & -- & -- & -- & -- \\
    \textcolor{gray}{SFD$^*$~\cite{sfd}} & \textcolor{gray}{0.26} & \textcolor{gray}{675M} & \textcolor{gray}{800} & \textcolor{gray}{1.06} & \textcolor{gray}{267.0} & \textcolor{gray}{0.78} & \textcolor{gray}{0.67} & \textcolor{gray}{--} & \textcolor{gray}{--} & \textcolor{gray}{--} & \textcolor{gray}{--} \\
    \midrule

    \multicolumn{12}{c}{\textit{\textbf{Long Period Training for Representation-Native DiT}}} \\
    \midrule
    SVG~\cite{svg} & 0.65 & 675M & 1400 & 1.92 & 264.9 & -- & -- & 3.36 & 181.2 & -- & -- \\
    VA-VAE~\cite{vavae} & 0.28 & 675M & 800 & 1.35 & 295.3 & 0.79 & 0.65 & 2.17 & 205.6 & 0.77 & 0.65 \\
    AlignTok~\cite{aligntok} & 0.26 & 675M & 800 & 1.37 & 293.6 & 0.79 & 0.65 & 2.04 & 206.2 & 0.76 & \textbf{0.67} \\
    Send-VAE (w. REPA)~\cite{sendvae} & 0.31 & 675M & 800 & 1.21 & 315.1 & 0.79 & 0.66 & 1.75 & 218.5 & 0.79 & 0.64 \\
    \textcolor{gray}{RAE (DiT-XL)$^*$~\cite{rae}} & \textcolor{gray}{0.57} & \textcolor{gray}{676M} & \textcolor{gray}{800} & \textcolor{gray}{1.41} & \textcolor{gray}{309.4} & \textcolor{gray}{0.80} & \textcolor{gray}{0.63} & \textcolor{gray}{1.87} & \textcolor{gray}{209.7} & \textcolor{gray}{0.80} & \textcolor{gray}{0.63} \\
    \textcolor{gray}{RAE (DiTDH-XL)$^*$~\cite{rae}} & \textcolor{gray}{0.57} & \textcolor{gray}{839M} & \textcolor{gray}{800} & \textcolor{gray}{1.13} & \textcolor{gray}{262.6} & \textcolor{gray}{0.78} & \textcolor{gray}{0.67} & \textcolor{gray}{1.51} & \textcolor{gray}{242.9} & \textcolor{gray}{0.79} & \textcolor{gray}{0.63} \\
    FAE~\cite{fae} & 0.68 & 675M & 800 & 1.29 & 268.0 & 0.80 & 0.64 & 1.48 & 239.8 & \textbf{0.81} & 0.63 \\
    VTP~\cite{vtp} & 0.36 & 675M & 600 & 1.11 & 279.5 & 0.79 & 0.67 & 1.85 & 232.3 & 0.79 & \textbf{0.67} \\
    \rowcolor{blue!8}
    \textbf{PAE (MAE)} & \textbf{0.23} & 675M & 800 & 1.78 & \textbf{368.0} & \textbf{0.81} & 0.65 & 2.83 & 189.4 & 0.75 & \textbf{0.67} \\
    \rowcolor{blue!8}
    \textbf{PAE (SigLIP2)} & 0.27 & 675M & 800 & 1.07 & 287.4 & 0.77 & \textbf{0.68} & 1.60 & 235.8 & 0.77 & 0.66 \\
    \rowcolor{blue!8}
    \textbf{PAE (DINOv3)} & 0.28 & 675M & 800 & 1.07 & 292.2 & 0.78 & 0.67 & 1.45 & \textbf{261.0} & 0.79 & 0.65 \\
    \rowcolor{blue!8}
    \textbf{PAE (DINOv2)} & 0.26 & 675M & 800 & \textbf{1.03} & 296.9 & 0.79 & 0.67 & \textbf{1.43} & 244.8 & 0.78 & 0.66 \\
    \bottomrule
    \end{tabular}}
    \vspace{-0.4cm}
\end{table*}

\begin{figure*}[t]
    \centering    
    \includegraphics[width=\textwidth]{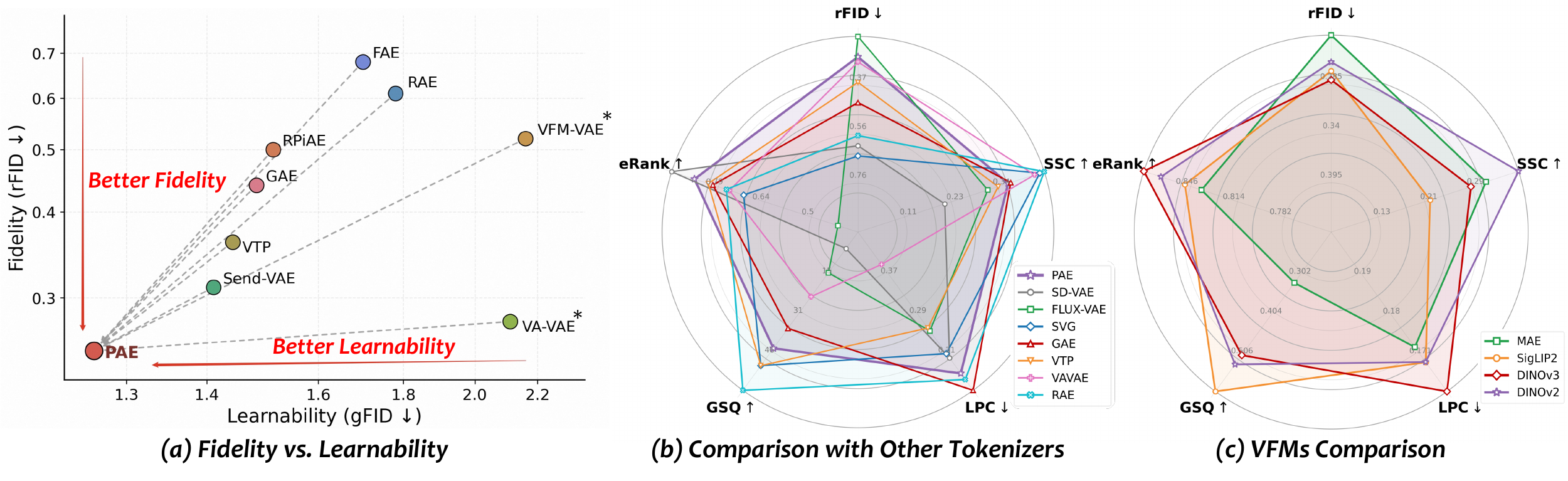}
    \vspace{-0.5cm}
    \caption{\textbf{Understanding PAE's fidelity--learnability advantage.} (a) Trade-off between reconstruction fidelity and downstream learnability across tokenizers. $*$ denotes generative performance measured at 64 training epochs. (b) Comparison of reconstruction, latent geometry, and utilization using rFID, SSC, LPC, GSQ, and eRank. (c) Profiles of PAE built on different VFM backbones.}
    \label{fig:metrics_comp}
    \vspace{-0.4cm}
\end{figure*}

\begin{figure*}[t]
    \centering    \includegraphics[width=\textwidth]{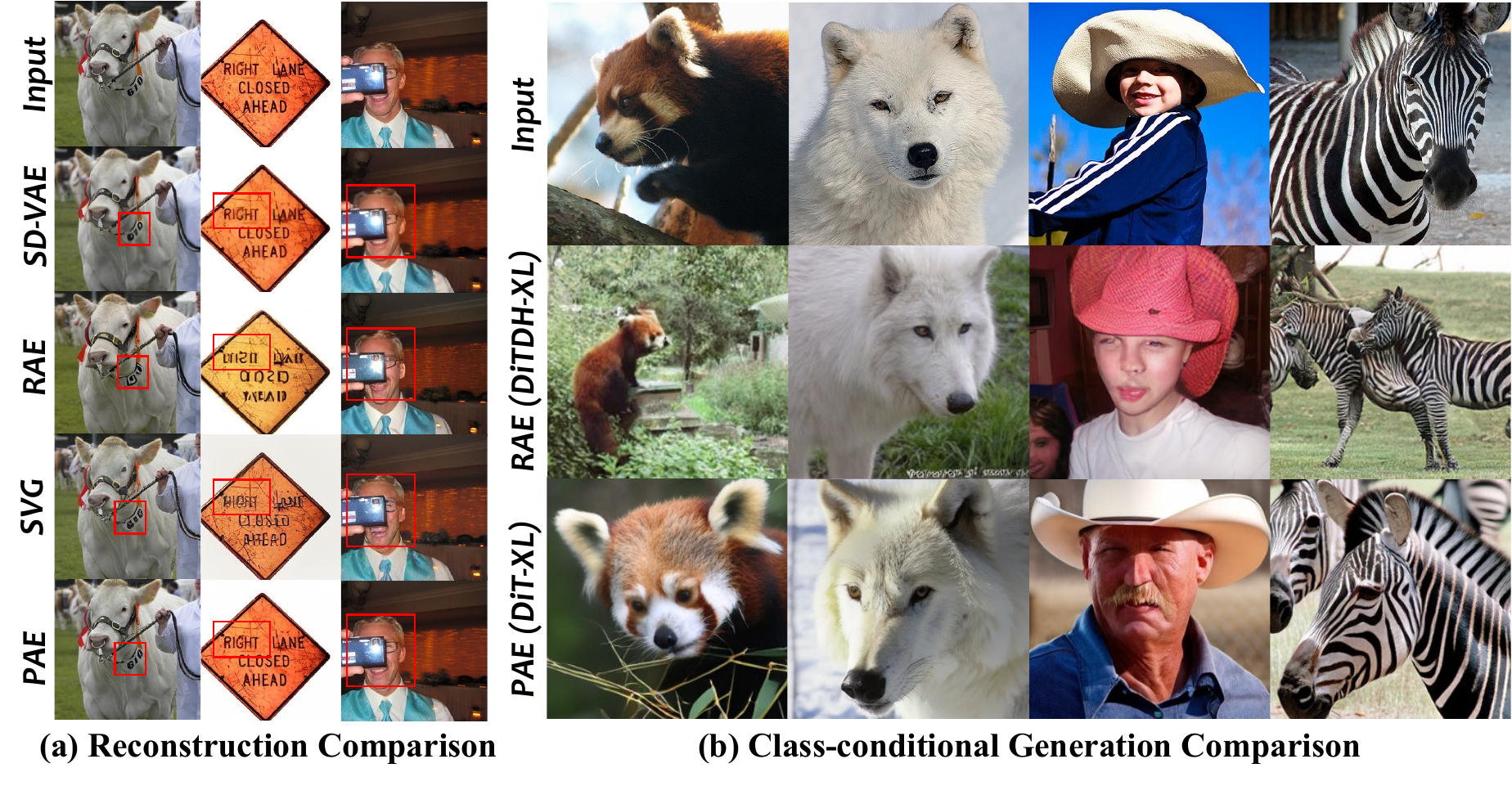}
    \vspace{-0.7cm}
    \caption{\textbf{Qualitative Comparison.} (a) \textit{Reconstruction:} PAE outperforms other tokenizers in reconstructing details (e.g., thin structures, text, and faces). (b) \textit{Generation:} $256\times256$ ImageNet samples from LightningDiT-XL/1 (80 epochs) demonstrating the high fidelity and coherence of PAE.}
    \label{fig:recon_gen}
    \vspace{-0.6cm}
\end{figure*}

\begin{figure*}[!t]
    \vspace{-0.4cm}
    \centering
    \begin{subfigure}[b]{0.31\textwidth}
        \centering
        \includegraphics[width=\linewidth]{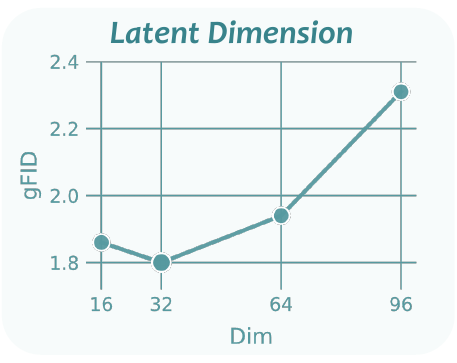}
        \caption{\textbf{Latent dimension.}}
        \label{fig:ab_dim}
    \end{subfigure}
    \hfill
    \begin{subfigure}[b]{0.31\textwidth}
        \centering
        \includegraphics[width=\linewidth]{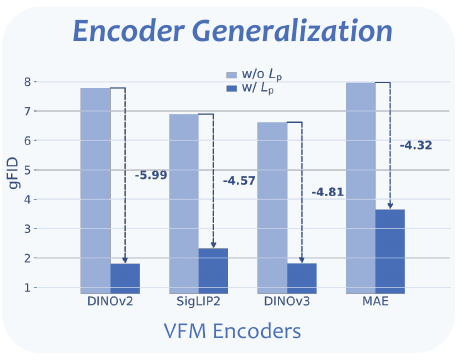}
        \caption{\textbf{Encoder generalization.}}
        \label{fig:ab_filter}
    \end{subfigure}
    \hfill
    \begin{subfigure}[b]{0.31\textwidth}
        \centering
        \includegraphics[width=\linewidth]{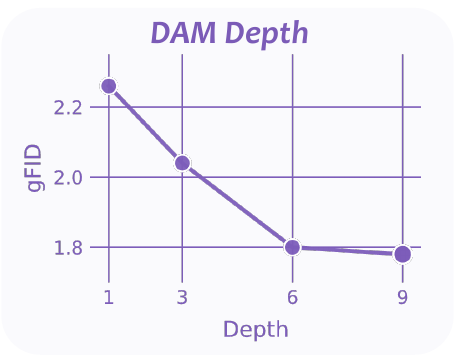}
        \caption{\textbf{DAM depth.}}
        \label{fig:ab_depth}
    \end{subfigure}
    \vspace{-0.1cm}
    \caption{\textbf{Generalization and design analysis.} PAE remains effective across teacher encoders and is stable under moderate changes in latent dimension and DAM depth.}
    \vspace{-0.6cm}
    \label{fig:sensitivity}
\end{figure*}

In this section, we evaluate \pae{} on ImageNet $256{\times}256$ and study the following questions:
\begin{itemize}[leftmargin=*,itemsep=0mm]
    \item \textbf{Q1: Model performance.} Can \pae{} improve downstream generation quality and convergence speed over strong latent-diffusion tokenizers? (\mbox{Tab.~\ref{tab:main_system}}, Fig.~\ref{fig:gen_vis}, Fig.~\ref{fig:metrics_comp}\textnormal{(a)}, Fig.~\ref{fig:recon_gen})
    \item \textbf{Q2: What explains \pae{}'s gains?} Do the geometry metrics and prior-alignment objectives explain \pae{}'s improved fidelity--learnability balance? (\mbox{Tab.~\ref{tab:factorial_prior}(a)}, Fig.~\ref{fig:metrics_comp}\textnormal{(b)(c)})
    \item \textbf{Q3: Ablation studies.} Are the proposed design choices effective, and does \pae{} remain robust across different encoders and moderate design changes? (\mbox{Tab.~\ref{tab:factorial_prior}(b)}, Tab.~\ref{tab:ablation_core}, Fig.~\ref{fig:sensitivity})
\end{itemize}

\textbf{Implementation Details.}
We consider multiple frozen representation encoders, including DINOv2-L~\cite{dinov2}, SigLIP2-SO400M~\cite{siglip2}, DINOv3-L~\cite{dinov3}, and MAE-L~\cite{mae}. Unless otherwise specified, all ablations use DINOv2-L. By default, the latent size is $16{\times}16{\times}32$, the Detail-aware Modulator (DAM) uses $K{=}6$ blocks, and the tokenizer is trained on ImageNet for 50 epochs with the joint objective in Eq.~\ref{eq:recon} and Eq.~\ref{eq:lp}. For downstream class-conditional generation, we train LightningDiT-XL on the same setup following VA-VAE~\cite{vavae}. Our experiments are conducted on NVIDIA A100 GPUs. More implementation details are provided in Appendix~\ref{app:implementation}.
\vspace{-0.1cm}

\textbf{Convergence Speed and Final Performance.}
Tab.~\ref{tab:main_system} reports both short-horizon convergence and final performance. At 80 generator epochs, PAE(DINOv2) reaches 1.27 guided gFID, outperforming strong representation-native baselines such as VTP (1.44) and GAE (1.48). It also surpasses RAE (DiTDH-XL), despite using fewer generator parameters (675M vs.\ 839M) and a simpler guidance strategy (CFG vs. AutoGuidance). This indicates that the latent space learned by \pae{} is easier for downstream diffusion to optimize, not merely better after long training.
With longer training, PAE(DINOv2) further reaches 1.03 guided gFID at 800 epochs, the best guided result among all compared methods, while also achieving strong unguided quality at 1.43 gFID. Fig.~\ref{fig:gen_vis} and Fig.~\ref{fig:recon_gen} show that these gains are accompanied by faithful reconstruction and high-quality image synthesis.
\vspace{-0.1cm}

\textbf{Why does \pae{} achieve a better fidelity--learnability balance?}
Fig.~\ref{fig:metrics_comp}\textnormal{(a)} shows that previous tokenizers typically trade reconstruction against learnability, whereas \pae{} achieves both. Fig.~\ref{fig:metrics_comp}\textnormal{(b)} suggests that this comes from a more balanced latent geometry, with strong spatial structure, local continuity, and global semantics. Fig.~\ref{fig:metrics_comp}\textnormal{(c)} further shows that DINO-based PAE is the most balanced and performs best, while SigLIP and MAE exhibit weaker geometry profiles on different dimensions. Together, these results suggest that \pae{} works best when reconstruction and the three primary geometry properties are jointly well balanced. More discussion is provided in Appendix~\ref{app:ablation}.
\vspace{-0.1cm}

\textbf{Effect of Prior-Alignment Objectives.}
Tab.~\ref{tab:factorial_prior}(a) ablates SSR, MCR, and SCR on top of the same baseline tokenizer, namely PAE without $\loss_p$. Each objective alone already yields a large gain over the baseline, and each one most strongly improves its intended geometry dimension: SSR improves SSC the most, MCR improves LPC the most, and SCR improves GSQ the most. The pairwise combinations further show complementarity, and the full model achieves the best overall result at 1.86 gFID and 210.8 IS. This confirms that \pae{} improves generation by jointly shaping structure, continuity, and semantics.
\vspace{-0.1cm}

\textbf{Impact of Refined Priors.}
Tab.~\ref{tab:factorial_prior}(b) isolates target construction under the same prior-alignment losses.
Refined VFM targets consistently improve SSC, GSQ, LPC, rFID, and gFID over raw targets, indicating cleaner and better bottleneck-matched supervision.
Still, this improvement is modest relative to the much larger gain from prior alignment in Tab.~\ref{tab:factorial_prior}(a), suggesting that PAE mainly benefits from the prior losses, with refinement serving as a complementary enhancement.
\vspace{-0.1cm}

\begin{table*}[t]
\centering
\caption{\textbf{Ablation study on prior alignment.} All ablations use 25 tokenizer epochs. (a) Each objective most strongly improves its intended dimension, and combining all three gives the best overall generation performance. (b) Refining the VFM targets further improves structure and semantics.}
\vspace{-0.1cm}
\label{tab:factorial_prior}
\small
\setlength{\tabcolsep}{5pt}
\renewcommand{\arraystretch}{1.08}

\begin{minipage}[t]{0.64\textwidth}
\centering
\textbf{(a) Prior alignment objectives}
\vspace{0.35em}

\begin{tabular}{ccc | cccccc}
\toprule
\textbf{SSR} & \textbf{MCR} & \textbf{SCR} & \textbf{SSC}$\uparrow$ & \textbf{LPC}$\downarrow$ & \textbf{GSQ}$\uparrow$ & \textbf{rFID}$\downarrow$ & \textbf{gFID}$\downarrow$ & \textbf{IS}$\uparrow$ \\
\midrule
\xmark     & \xmark     & \xmark     & 0.18 & 0.320 & 0.19 & 0.24 & 7.18 & 117.2 \\
\checkmark & \xmark     & \xmark     & 0.29 & 0.296 & 0.26 & 0.25 & 2.74 & 161.8 \\
\xmark     & \checkmark & \xmark     & 0.23 & 0.221 & 0.24 & 0.26 & 2.53 & 173.6 \\
\xmark     & \xmark     & \checkmark & 0.21 & 0.286 & 0.39 & 0.25 & 2.63 & 168.4 \\
\checkmark & \checkmark & \xmark     & 0.33 & 0.187 & 0.33 & 0.26 & 2.02 & 194.6 \\
\checkmark & \xmark     & \checkmark & 0.31 & 0.258 & 0.46 & 0.26 & 2.10 & 188.9 \\
\xmark     & \checkmark & \checkmark & 0.24 & 0.176 & 0.45 & 0.27 & 2.08 & 191.3 \\
\rowcolor{blue!8}\checkmark & \checkmark & \checkmark & 0.35 & 0.170 & 0.50 & 0.26 & 1.86 & 210.8 \\
\bottomrule
\end{tabular}

\end{minipage}
\hfill
\begin{minipage}[t]{0.30\textwidth}
\centering

\textbf{(b) VFM priors}
\vspace{0.35em}

\begin{tabular}{l|cc}
\toprule
\textbf{Metric} & \textbf{Raw} & \textbf{Refined} \\
\midrule
\addlinespace[0.8ex]
SSC$\uparrow$    & 0.33 & 0.35 \\
\addlinespace[1.9ex]
LPC$\downarrow$  & 0.171 & 0.170 \\
\addlinespace[1.9ex]
GSQ$\uparrow$    & 0.48 & 0.50 \\
\addlinespace[1.9ex]
rFID$\downarrow$ & 0.27 & 0.26 \\
\addlinespace[1.9ex]
gFID$\downarrow$ & 1.95 & 1.86 \\
\addlinespace[0.2ex]
\bottomrule
\end{tabular}
\end{minipage}
\end{table*}

\textbf{Core Design Ablations.}
Tab.~\ref{tab:ablation_core}(a) compares our prior-alignment design against several generic latent regularization baselines, including a weak KL penalty and a lightweight diffusion-loss regularizer; detailed settings are provided in Appendix~\ref{app:core_ablation_settings}. Generic regularizers help, but remain much weaker than our manifold-targeted alignment (5.17 / 4.22 vs.\ 1.80 gFID), indicating that the gain comes from regularizing the latent properties rather than from regularization alone. Tab.~\ref{tab:ablation_core}(b) shows that DAM outperforms direct finetuning and simple residual fusion, supporting controlled detail injection. Tab.~\ref{tab:ablation_core}(c) shows that full token-level SCR supervision performs best, confirming the importance of preserving dense semantic directions.
\vspace{-0.1cm}

\textbf{Sensitivity and Generalization.}
Fig.~\ref{fig:ab_dim} shows that performance peaks at a moderate latent dimension, indicating that \pae{} benefits from sufficient but not excessive capacity. Fig.~\ref{fig:ab_filter} shows that the gain from $\loss_p$ is consistent across DINOv2, SigLIP2, DINOv3, and MAE. Fig.~\ref{fig:ab_depth} shows that DAM depth helps up to a moderate range and then saturates, indicating stable behavior under reasonable design changes.
\vspace{-0.1cm}

\textbf{More Ablation Studies.}
Additional results, including full encoder comparisons, diagnostic correlations, few-step sampling, latent robustness, and more visualizations, are provided in Appendix~\ref{app:ablation}.
\vspace{-0.6cm}

\begin{table*}[t]
\vspace{-0.6cm}
\centering
\caption{\textbf{Ablation study on core design choices based on PAE (DINOv2).}}
\vspace{-0.1cm}
\label{tab:ablation_core}
\small
\setlength{\tabcolsep}{4pt}
\renewcommand{\arraystretch}{1.0}

\begin{minipage}[t]{0.31\textwidth}
\centering
\textbf{(a) Regularization Strategy}

\vspace{0.3em}
\begin{tabular}{lcc}
\toprule
\textbf{Method} & \textbf{gFID}$\downarrow$ & \textbf{IS}$\uparrow$ \\
\midrule
Baseline     & 7.79 & 117.2 \\
KL Reg       & 5.17 & 132.4 \\
Diff Reg     & 4.22 & 148.3 \\
\rowcolor{blue!8}\textbf{Ours} & \textbf{1.80} & \textbf{218.3} \\
\bottomrule
\end{tabular}
\end{minipage}
\hfill
\begin{minipage}[t]{0.31\textwidth}
\centering
\textbf{(b) Detail Injection}

\vspace{0.3em}
\begin{tabular}{lcc}
\toprule
\textbf{Method} & \textbf{gFID}$\downarrow$ & \textbf{IS}$\uparrow$ \\
\midrule
Finetuning   & 2.13 & 198.8 \\
Res. Add     & 2.46 & 187.9 \\
Res. Concat  & 2.38 & 192.5 \\
\rowcolor{blue!8}\textbf{DAM} & \textbf{1.80} & \textbf{218.3} \\
\bottomrule
\end{tabular}
\end{minipage}
\hfill
\begin{minipage}[t]{0.31\textwidth}
\centering
\textbf{(c) SCR Semantic Supervision}

\vspace{0.3em}
\begin{tabular}{lcc}
\toprule
\textbf{Target} & \textbf{gFID}$\downarrow$ & \textbf{IS}$\uparrow$ \\
\midrule
\addlinespace[4pt]
Pooling Token  & 2.14 & 201.2 \\
\addlinespace[4pt]
Feature Tokens & 1.87 & 206.8 \\
\addlinespace[4pt]
\rowcolor{blue!8}\textbf{Full Tokens} & \textbf{1.80} & \textbf{218.3} \\
\addlinespace[1pt]
\bottomrule
\end{tabular}
\end{minipage}
\end{table*}

%% file: sections/conclusion.tex
\section{Conclusion}
\label{sec:conclusion}
\vspace{-0.1cm}

In this paper, we propose \textit{\textbf{P}rior-\textbf{A}ligned Auto\textbf{e}ncoders} (\textbf{PAE}), a tokenizer framework for improving latent diffusion through explicit latent-manifold shaping. Through pilot analysis and large-scale experiments, we show that reconstruction quality alone is insufficient to explain tokenizer effectiveness, and that stronger generation is more closely associated with latent spaces that preserve instance-level structure, local continuity, and global semantics. To this end, PAE introduces prior-alignment objectives that explicitly regularize these three properties during tokenizer learning. Extensive experiments on ImageNet $256{\times}256$ demonstrate that PAE consistently improves both generation quality and convergence speed, reaching comparable quality to RAE with substantially fewer training epochs under the same LightningDiT setup and achieving 1.03 gFID in long training.

%% file: sections/appendix.tex
\section{Extended Related Works}
\label{app:ex_related_work}

Research on visual tokenizers for downstream generation has expanded rapidly, covering reconstruction-oriented autoencoders, representation-based tokenizers, unified tokenizers for understanding and generation, and tokenizer-side regularization tailored to diffusion. Despite their diversity in architecture and objective, many of these methods can be viewed as improving the \emph{diffusability} or \emph{learnability} of the induced latent space. We organize this literature through the lens of \emph{latent manifold organization}, and highlight two directions most relevant to our work: \emph{representation-centric} methods, which inherit stronger pretrained visual priors, and \emph{spectral-/structure-centric} methods, which reshape the spatial, frequency, or channel organization of latent codes.

\paragraph{Broader tokenizer landscape.}
Visual tokenizers were originally developed from the perspective of compression and reconstruction, including classical AEs~\cite{hinton2006reducing}, VAEs~\cite{kingma2013auto}, VQ-VAEs~\cite{razavi2019generating}, and VQGAN~\cite{esser2021taming}. Subsequent work improved quantization efficiency, codebook usage, scalability, and adaptive token allocation, including BSQ~\cite{bsq}, IBQ~\cite{ibq}, DC-AE~\cite{dcae}, MAE-Tok~\cite{maetok}, CAT~\cite{cat}, ElasticTok~\cite{elastictok}, and FlexTok~\cite{flextok}. Although not all of these methods target diffusion explicitly, they establish the key trade-offs between reconstruction fidelity, compression ratio, latent capacity, and generative learnability. In particular, CRT~\cite{crt} and VA-VAE~\cite{vavae} show that stronger reconstruction does not necessarily imply better generation, motivating tokenizer analysis beyond pixel fidelity alone.

\paragraph{(1) Representation-centric approaches to diffusability.}
A major direction is to improve generation by leveraging pretrained visual foundation models as tokenizers, encoders, or teachers. Some methods directly build autoencoders or latent generators on top of pretrained representation encoders, including RAE~\cite{rae}, Scale-RAE~\cite{scale_rae}, FlatDINO~\cite{flatdino}, LV-RAE~\cite{lvrae}, DINO-SAE~\cite{dinosae}, FAE~\cite{fae}, SVG~\cite{svg}, VFMTok~\cite{vfmtok}, RepTok~\cite{reptok}, and VFM-VAE~\cite{vfmvae}. Others use frozen VFM features as alignment targets or supervision during tokenizer training, such as GAE~\cite{gae}, AlignTok~\cite{aligntok}, REPA-E~\cite{repa-e}, VA-VAE~\cite{vavae}, and PS-VAE~\cite{psvae}. Closely related are unified tokenizers that aim to support both understanding and generation while preserving pretrained semantics, including UniFlow~\cite{uniflow}, UniTok~\cite{unitok}, UniLIP~\cite{unilip}, and TokenFlow~\cite{tokenflow}. A related direction is VTP~\cite{vtp}, which learns strong semantics through large-scale tokenizer pretraining rather than explicit frozen-teacher alignment.
The common intuition is that pretrained or representation-rich features are semantically stronger, often smoother and more spatially regular, and therefore easier for downstream generators to model than purely reconstruction-oriented latents. In this sense, these methods can be viewed as implicitly improving latent diffusability through stronger representation priors. However, most of them primarily emphasize representation inheritance rather than explicit latent organization for diffusion. Methods that expose the generator more directly to pretrained encoder features, such as RAE~\cite{rae}, FAE~\cite{fae}, and SVG~\cite{svg}, inherit strong semantics but often struggle with faithful pixel reconstruction because pretrained encoders are not optimized for fine-grained detail. Teacher-alignment methods such as GAE~\cite{gae}, AlignTok~\cite{aligntok}, REPA-E~\cite{repa-e}, and VA-VAE~\cite{vavae} mitigate this issue by learning dedicated tokenizers, but typically align to raw frozen features that may remain high-dimensional, bottleneck-mismatched, and spatially imperfect at tokenizer resolution. VTP~\cite{vtp} avoids an explicit frozen teacher, yet still learns the resulting geometry only implicitly. In contrast, our method refines VFM features into tokenizer-compatible priors and uses them to explicitly shape spatial structure, local continuity, and global semantic organization.

\paragraph{(2) Spectral and structure-centric approaches to diffusability.}
Another line of work improves tokenizer diffusability by reshaping latent organization from the viewpoint of spectral bias or structural regularity. On the spectral side, diffusion models under Gaussian noising often exhibit a coarse-to-fine or approximately spectral-autoregressive generation process, as discussed in spectral autoregression analyses~\cite{rissanen2023generativemodellinginverseheat, ser} and analytical studies of diffusion spectral bias~\cite{wang2026analyticaltheoryspectralbias}. This motivates tokenizer-side methods that suppress excessive high-frequency content, strengthen low-frequency bias, or reorganize the latent spectrum to better match downstream diffusion. SER~\cite{ser}, EQ-VAE~\cite{eqvae}, and Denoising-VAE~\cite{denoisingvae} reduce decoder reliance on high-frequency latent signals through low-pass-consistent, scale-equivariant, or spectral denoising objectives, while UAE~\cite{uae} decomposes latent features into frequency bands, anchoring semantics in lower-frequency components and treating higher-frequency components as residual detail.
On the structural side, several works improve diffusability by encouraging smoother spatial organization, stronger local correlation, or lower channel redundancy. SSVAE~\cite{ssvae} studies both low-frequency spatio-temporal bias and few-mode-biased channel statistics, and introduces local correlation regularization and latent masked reconstruction to promote smoother local organization and more concentrated channel usage. DCAE1.5~\cite{dcae15} similarly uses channel masking and redundancy-reduction strategies that reduce the burden of modeling redundant channels. Although motivated differently, these methods point to a common conclusion: diffusion-friendly latents depend not only on semantics or reconstruction quality, but also on how information is organized across frequency, space, and channel dimensions.

\paragraph{Our manifold-centered perspective.}
These lines of work provide complementary evidence that downstream diffusion depends on latent organization. Representation-centric methods highlight the value of stronger semantic priors~\cite{rae, fae, svg, gae}, while spectral and structure-centric methods emphasize lower-frequency organization, local smoothness, and reduced redundancy~\cite{ser, denoisingvae, uae, ssvae, dcae15}. Our perspective is to unify these observations through \emph{latent manifold organization}. From this viewpoint, tokenizer quality is determined not only by reconstruction fidelity or representation strength, but also by whether the induced latent space exhibits geometry that makes diffusion learning easier. Concretely, we focus on three complementary properties associated with downstream generation quality in our analysis: coherent instance-level spatial structure, local manifold continuity, and global semantic organization.
This perspective also offers a common language for prior methods. Representation-based approaches such as RAE~\cite{rae}, SVG~\cite{svg}, and GAE~\cite{gae} can be interpreted as improving semantic organization or spatial regularity of the latent manifold. Spectral methods such as SER~\cite{ser}, Denoising-VAE~\cite{denoisingvae}, and UAE~\cite{uae} can be interpreted as biasing the manifold toward smoother coarse structure and reduced high-frequency noise. Structural methods such as SSVAE~\cite{ssvae} and DCAE1.5~\cite{dcae15} can be interpreted as simplifying local geometry and channel organization. Building on these insights, our framework refines frozen VFM features into tokenizer-compatible priors and explicitly regularizes the latent space along the three manifold dimensions above.

\section{Latent Manifold Geometry Metrics}
\label{app:metrics_theory}

In this appendix, we formalize three complementary properties of tokenizer-induced latent geometry and use them as \emph{empirical diagnostics} for analyzing when a latent space is diffusion-friendly beyond reconstruction quality alone. Our goal is not to claim that these metrics are formally derived complexity measures for diffusion training. Instead, they operationalize three latent properties that are intuitively relevant to downstream generation and are supported in our paper by controlled empirical comparisons and correlation analysis: instance-level spatial structure, local continuity, and semantic neighborhood quality. We additionally report an effective-rank diagnostic to characterize latent utilization.

\subsection{Metric Definitions}
\label{app:metric_definitions}

We define three primary latent geometry metrics that capture instance-level spatial structure, local perceptual continuity, and semantic neighborhood quality, respectively. In addition, we report a supplementary latent-complexity diagnostic based on effective rank.

\begin{figure}[ht]
    \centering
    \includegraphics[width=\textwidth]{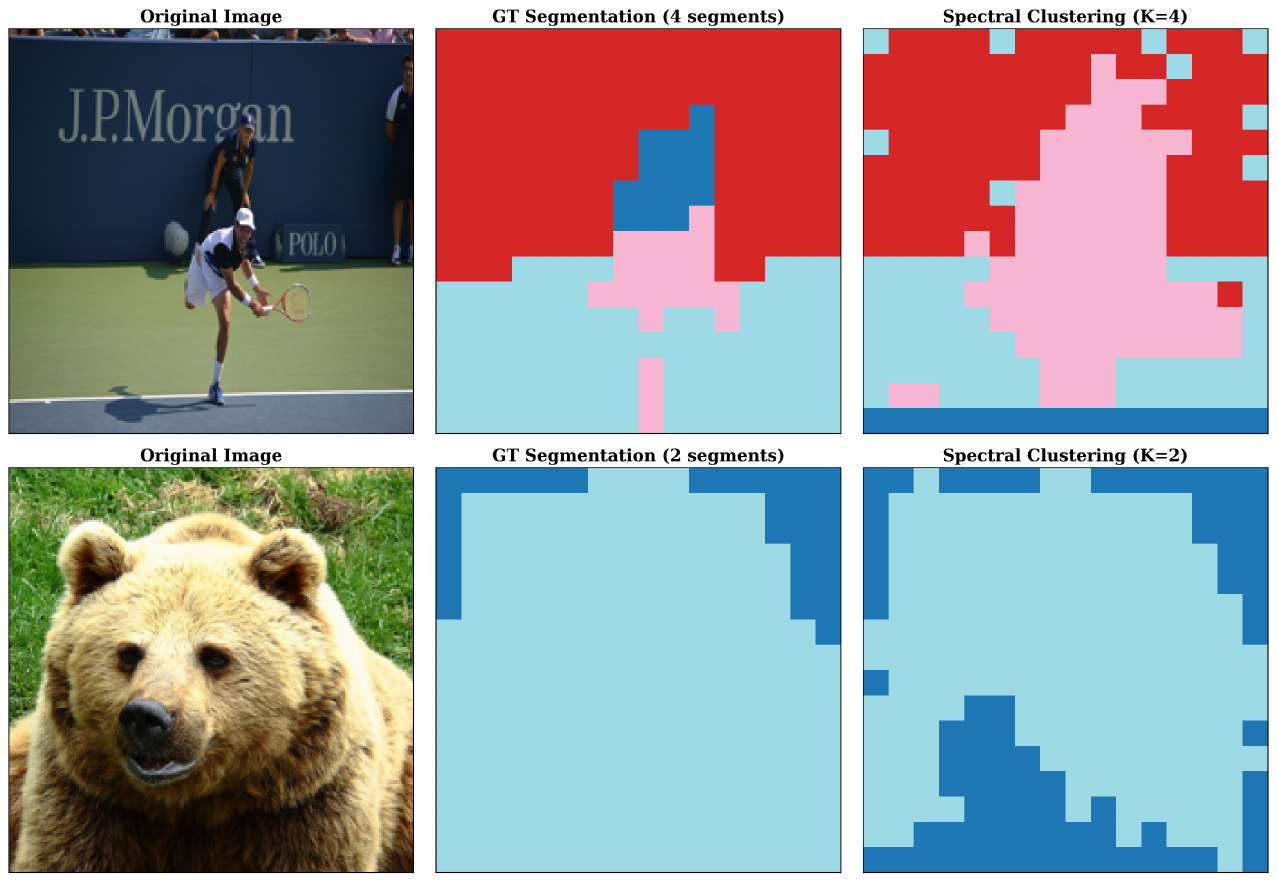}
    \caption{
    \textbf{Illustration of Spatial Structure Coherence (SSC).}
    For each image, we construct a latent-token affinity graph from the tokenizer output, perform spectral clustering on the token graph, and compare the resulting token partition with object-aware panoptic labels projected to latent resolution.
    Higher SSC indicates better alignment between latent token grouping and object-level spatial structure.
    }
    \label{fig:ssc_vis}
\end{figure}

\paragraph{Metric I: Spatial Structure Coherence (SSC).}
SSC measures whether the spatial organization of latent tokens preserves object-aware structure within each instance.

For a latent tensor $\mathbf{Z}\in\mathbb{R}^{C\times H\times W}$, let $N=HW$ and reshape $\mathbf{Z}$ into token vectors
\[
\{\mathbf{z}_i\}_{i=1}^N,\qquad \mathbf{z}_i\in\mathbb{R}^C.
\]
We define the latent-token affinity matrix
\begin{equation}
A_{ij}
=
\exp\!\left(
\frac{\langle \hat{\mathbf{z}}_i,\hat{\mathbf{z}}_j\rangle}{\sigma}
\right),
\qquad
\hat{\mathbf{z}}_i=\frac{\mathbf{z}_i}{\|\mathbf{z}_i\|_2},
\qquad
A_{ii}=0,
\label{eq:ssc_affinity_final}
\end{equation}
where $\sigma>0$ is a temperature parameter.

Applying normalized spectral clustering~\cite{spectral_clustering} to $A$ produces predicted token labels
\[
\hat{\mathbf{y}}=(\hat y_1,\dots,\hat y_N).
\]
Let
\[
\mathbf{y}=(y_1,\dots,y_N)
\]
denote object-level ground-truth labels projected to latent resolution. We obtain $\mathbf{y}$ from COCO Panoptic Val 2017~\cite{coco2017} annotations by applying the same spatial transformation as the image and downsampling the panoptic mask to latent resolution via majority vote. In practice, the number of clusters is set to the number of object segments present in the projected ground-truth mask for each sample. We then define Spatial Structure Coherence as the geometric-mean normalized mutual information
\begin{equation}
\mathrm{SSC}(\mathbf{y},\hat{\mathbf{y}})
=
\frac{I(\mathbf{y};\hat{\mathbf{y}})}
{\sqrt{H(\mathbf{y})H(\hat{\mathbf{y}})}},
\label{eq:ssc_metric_final}
\end{equation}
where $I(\cdot;\cdot)$ is mutual information and $H(\cdot)$ is Shannon entropy. Higher SSC indicates that latent token grouping better preserves object-aware spatial structure. An illustration is shown in Fig.~\ref{fig:ssc_vis}.

\paragraph{Metric II: Local Perceptual Continuity (LPC).}
To measure whether local latent perturbations induce smooth and stable perceptual changes after decoding, we define a perceptual continuity metric based on LPIPS distance.

Let $\mathbf{z}=E(\mathbf{x})$ be the flattened latent code of an image $\mathbf{x}$, and let $\mathbf{u}\sim \mathrm{Unif}(\mathbb{S}^{d-1})$ be a random unit perturbation direction in latent space. Let $d_{\mathrm{LPIPS}}(\cdot,\cdot)$ denote the LPIPS perceptual distance. For a perturbation scale $\epsilon>0$, we define the single-scale local perceptual continuity as
\begin{equation}
\mathrm{LPC}_{\epsilon}
=
\mathbb{E}_{\mathbf{x},\mathbf{u}}
\left[
\frac{
d_{\mathrm{LPIPS}}\!\left(D(\mathbf{z}+\epsilon\mathbf{u}),D(\mathbf{z})\right)
+
d_{\mathrm{LPIPS}}\!\left(D(\mathbf{z}-\epsilon\mathbf{u}),D(\mathbf{z})\right)
}{2}
\right].
\label{eq:lpc_single_scale}
\end{equation}
Smaller $\mathrm{LPC}_{\epsilon}$ indicates that the decoded perceptual representation changes less under local latent perturbations.

In practice, we evaluate LPC over a finite set of relative perturbation scales
\[
\epsilon_s = \rho_s \|\mathbf{z}\|_2,
\qquad
\rho_s \in \mathcal R,
\]
where $\mathcal R=\{0.1,0.5,1.0,2.0\}$ in our implementation. We include moderately larger perturbation scales only to improve robustness of the diagnostic beyond infinitesimal neighborhoods, while assigning larger weights to smaller scales. We then define the multi-scale LPC as a weighted average
\begin{equation}
\mathrm{LPC}
=
\sum_{s=1}^{|\mathcal R|} w_s \,\mathrm{LPC}_{\epsilon_s},
\qquad
w_s
=
\frac{\rho_s^{-1}}{\sum_{r=1}^{|\mathcal R|}\rho_r^{-1}},
\label{eq:lpc_weighted}
\end{equation}
so that smaller perturbation scales receive larger weights. This construction emphasizes the local continuity of the decoder-induced perceptual neighborhood while remaining numerically stable in practice.

\paragraph{Interpretation.}
LPC measures perceptual stability of the decoder under local latent perturbations. Unlike a purely differential curvature quantity, LPC is defined directly through decoded perceptual distances and therefore reflects the operational behavior of the tokenizer--decoder pair. A smaller LPC indicates that nearby latent points remain perceptually close after decoding, which is desirable for diffusion-style local prediction. At the same time, LPC should be interpreted jointly with reconstruction and semantic metrics, since an overly insensitive decoder can also yield artificially small local perceptual changes.

\paragraph{Metric III: Global Semantic Quality (GSQ).}
GSQ measures global semantic organization through local nearest-neighbor purity in pooled latent space. Concretely, it tests whether the most similar latent neighbor of each sample belongs to the same semantic class.

For each image $\mathbf{x}_i$, let
\begin{equation}
\mathbf{f}_i
=
\mathrm{GAP}(\mathbf{Z}_i)
=
\frac{1}{HW}\sum_{h=1}^H\sum_{w=1}^W \mathbf{Z}_i[:,h,w]
\in\mathbb{R}^C
\label{eq:gsq_gap}
\end{equation}
be the globally pooled latent feature. We mean-center and $\ell_2$-normalize these features:
\begin{equation}
\bar{\mathbf{f}}
=
\frac{1}{N_{\mathcal C}}\sum_{i=1}^{N_{\mathcal C}} \mathbf{f}_i,
\qquad
\tilde{\mathbf{f}}_i
=
\frac{\mathbf{f}_i-\bar{\mathbf{f}}}{\|\mathbf{f}_i-\bar{\mathbf{f}}\|_2},
\label{eq:gsq_norm_final}
\end{equation}
where the average is taken over the sampled evaluation subset.

Let $\mathcal C$ be a random subset of $K'=100$ ImageNet~\cite{imagenet} classes sampled for computational efficiency, and let $N_{\mathcal C}$ denote the total number of images in this subset. In practice, we report mean and standard deviation across five random class subsets. For each sample $i$, define the index of its nearest latent neighbor under cosine similarity as
\begin{equation}
j^\star(i)
=
\arg\max_{j\neq i}\;
\langle \tilde{\mathbf{f}}_i,\tilde{\mathbf{f}}_j\rangle .
\label{eq:gsq_neighbor}
\end{equation}
We then define the \emph{Global Semantic Quality} as
\begin{equation}
\mathrm{GSQ}
=
\frac{1}{N_{\mathcal C}}
\sum_{i=1}^{N_{\mathcal C}}
\mathbf{1}\!\left[y_{j^\star(i)} = y_i\right],
\label{eq:gsq_metric_final}
\end{equation}
where $y_i$ is the class label of sample $i$ and $\mathbf{1}[\cdot]$ denotes the indicator function. A larger GSQ indicates that local semantic neighborhoods in latent space are purer and more class-consistent.

\paragraph{Interpretation.}
GSQ is not a class-centroid compactness measure; rather, it evaluates whether semantic nearest neighbors in latent space are label-consistent. This makes it especially suitable for retrieval-style or representation-rich tokenizers whose class manifolds may be locally pure without being globally unimodal.

\paragraph{Supplementary Diagnostic: Effective Rank Ratio (eRank).}
In addition to the three primary geometry metrics above, we report a supplementary latent-complexity diagnostic that measures how fully a latent representation utilizes its channel capacity.

Let $\mathbf{F}\in\mathbb{R}^{N\times C}$ be the feature matrix formed by globally pooled latent features over a dataset of $N$ images, after mean-centering. Let $\{\sigma_i\}_{i=1}^{C}$ be the singular values of $\mathbf{F}$, and define the normalized singular-value distribution
\[
\bar{\sigma}_i
=
\frac{\sigma_i}{\sum_j \sigma_j}.
\]
Following the entropy-based effective rank of Roy and Vetterli, we define
\begin{equation}
\mathrm{erank}(\mathbf{F})
=
\exp\!\left(
-\sum_{i=1}^{C}\bar{\sigma}_i\log \bar{\sigma}_i
\right).
\label{eq:erank_raw}
\end{equation}
We further normalize by the channel dimension and define
\begin{equation}
\mathrm{eRank}
=
\frac{\mathrm{erank}(\mathbf{F})}{C}.
\label{eq:erank_ratio}
\end{equation}
A larger eRank indicates that latent channels are utilized more evenly, whereas a smaller eRank suggests concentration in a few dominant directions. We use eRank only as a supplementary diagnostic of latent utilization rather than as a primary geometry objective.

\subsection{Why These Metrics Matter for Diffusion Learning}
\label{app:theory_connection}

The proposed metrics are intended as \emph{empirical geometry diagnostics} rather than formally derived complexity measures for diffusion training. This subsection therefore provides geometric intuition for why SSC, LPC, and GSQ are relevant to downstream DiT learning, without claiming a strict causal or theorem-level relationship. The central idea is that downstream diffusion becomes easier when the tokenizer induces a latent space whose spatial relations are more coherent, whose local neighborhoods are more stable, and whose semantic neighborhoods are less mixed.

\paragraph{SSC and structured token interactions.}
SSC measures whether latent tokens group according to object-aware spatial structure. In transformer-based generators, this property is relevant because self-attention operates over token-token relations rather than over pixels directly. When the tokenizer produces spatially coherent token organization, the induced token graph is typically less fragmented: tokens belonging to the same object or region are more likely to remain mutually consistent, and long-range relational corrections caused by spatially incoherent tokenization become less necessary. Under this interpretation, higher SSC indicates that the latent representation is more compatible with structured token interactions, which can make self-attention-based diffusion models easier to optimize.

\paragraph{LPC and local neighborhood stability.}
LPC measures how much decoded images drift perceptually under local latent perturbations. If nearby latent codes decode to perceptually similar outputs, then the encoder-decoder pair induces a locally stable neighborhood in the operational sense most relevant to generation. This is useful for diffusion learning because flow matching and denoising are local prediction problems: nearby points in latent space should ideally correspond to nearby prediction targets. Under this interpretation, smaller LPC suggests that local neighborhoods are more regular and that the target field varies more smoothly in practice, which can improve optimization stability.

\paragraph{GSQ and semantic neighborhood purity.}
GSQ measures whether nearest latent neighbors are also semantic neighbors. This property is useful because local semantic purity reduces class mixing in latent neighborhoods, which in turn makes the prediction target less heterogeneous in a neighborhood. In class-conditional generation, such organization is especially relevant, since the generator benefits from a latent space in which semantically related samples occupy more consistent local regions. GSQ should therefore be interpreted as a practical indicator of semantic neighborhood quality rather than as a direct estimate of any formal ambiguity quantity.

\paragraph{Unified perspective.}
Taken together, the three metrics provide complementary empirical views of diffusion-friendly latent organization:
\begin{itemize}
    \item \textbf{SSC} characterizes whether token interactions respect coherent spatial structure;
    \item \textbf{LPC} characterizes whether local latent neighborhoods remain stable under decoding;
    \item \textbf{GSQ} characterizes whether semantic neighborhoods are locally pure and well organized.
\end{itemize}
These quantities are complementary rather than redundant: they describe spatial organization, local continuity, and semantic neighborhood quality at different levels of latent geometry. In our paper, their usefulness is supported primarily by controlled comparisons and correlation analysis with downstream generation quality.

\paragraph{Role of eRank.}
Unlike SSC, LPC, and GSQ, the effective rank ratio eRank is not treated as a primary geometry metric and is not tied to a specific regularization objective. Instead, it serves as a supplementary diagnostic of latent utilization and effective degrees of freedom. In particular, eRank helps explain why some high-dimensional tokenizers may retain strong semantic or local geometric properties yet still remain harder to model generatively under a fixed DiT capacity budget.

\section{More Implementation Details}
\label{app:implementation}

\subsection{Main Experiment Configurations}
Table~\ref{tab:pae_config} summarizes the main experimental configurations for PAE and the latent diffusion generator. Unless otherwise specified, all experiments are conducted on ImageNet $256\times256$. For generator training, we follow the standard LightningDiT-XL setup used in prior representation-native autoencoder works such as VA-VAE~\cite{vavae} and GAE~\cite{gae}. Following prior LightningDiT practice, we use the standard 80 epoch and 800 epoch generator configurations, which differ in QK-Norm as commonly adopted in this benchmark setting.

\begin{table*}[ht]
\centering
\caption{\textbf{Configurations of PAE experiments.} All experiments are conducted on ImageNet $256\times256$ under the same generator setup unless otherwise specified.}
\renewcommand{\arraystretch}{1.2}
\setlength{\tabcolsep}{7pt}
\begin{tabular}{l|cccc}
\toprule
\rowcolor{gray!20}\multicolumn{5}{c}{\textbf{Tokenizer Architecture \& Training Setting}} \\
representation backbone & MAE-L & SigLIP2-SO400M & DINOv3-L & DINOv2-L \\
input image size & 256 & 256 & 256 & 224 \\
DAM hidden dim  & 1024 & 1152 & 1024 & 1024 \\
DAM depth & \multicolumn{4}{c}{6} \\
latent size & \multicolumn{4}{c}{$16\times16$} \\
latent dimension & \multicolumn{4}{c}{32} \\
decoder size & \multicolumn{4}{c}{ViT-L} \\
training epochs & \multicolumn{4}{c}{50} \\
warmup epochs & \multicolumn{4}{c}{1} \\
batch size & \multicolumn{4}{c}{512} \\
optimizer & \multicolumn{4}{c}{AdamW, $\beta_1$, $\beta_2$ = 0.9, 0.98} \\
learning rate & \multicolumn{4}{c}{2e-4 cosine decay to 2e-5} \\
loss weights & \multicolumn{4}{c}{$\lambda_{\mathrm{lpips}}{=}1.0$, $\lambda_{\mathrm{gan}}{=}0.5$,  $\lambda_{\mathrm{ssr}}{=}0.2$, $\lambda_{\mathrm{mcr}}{=}0.5$, $\lambda_{\mathrm{scr}}{=}1.0$} \\
MCR perturbation & \multicolumn{4}{c}{small: $42.5^\circ$, large: $85^\circ$ on latent sphere} \\
discriminator architecture & \multicolumn{4}{c}{DINO-S/8} \\
discriminator start epoch & \multicolumn{4}{c}{12} \\
discriminator update start epoch & \multicolumn{4}{c}{15} \\
discriminator optimizer & \multicolumn{4}{c}{AdamW, $\beta_1{=}0.9$, $\beta_2{=}0.98$, lr $=2\times10^{-4}$} \\
discriminator lr & \multicolumn{4}{c}{2e-4 cosine decay to 2e-5} \\

\midrule
\rowcolor{gray!20}\multicolumn{5}{c}{\textbf{LDM Architecture \& Training Setting}} \\
generator backbone & \multicolumn{4}{c}{LightningDiT-XL/1} \\
hidden dim & \multicolumn{4}{c}{1152} \\
depth & \multicolumn{4}{c}{28} \\
latent shape & \multicolumn{4}{c}{$16\times16\times32$} \\
QK Norm & \multicolumn{4}{c}{False (for 80 ep), True (for 800 ep)} \\
training epochs & \multicolumn{4}{c}{80 (\textit{Convergence Efficiency}), 800 (\textit{Final Performance})} \\
optimizer & \multicolumn{4}{c}{AdamW, $\beta_1$ = 0.9, $\beta_2$ = 0.95} \\
batch size & \multicolumn{4}{c}{1024} \\
learning rate & \multicolumn{4}{c}{2e-4} \\
learning rate schedule & \multicolumn{4}{c}{constant} \\
training time shift & \multicolumn{4}{c}{0.7} \\
\bottomrule
\end{tabular}
\label{tab:pae_config}
\end{table*}

For MCR, perturbations are applied in the RMS-normalized sphere-like latent space. Specifically, we sample a random normalized direction and use two perturbation levels, with maximum angular deviations of $42.5^\circ$ and $85^\circ$ for the small and large perturbations, respectively.

\subsection{Refining VFM Prior}
\label{app:stage0}

Before training the final PAE tokenizer, we first construct refined VFM-derived supervision targets through a separate target-construction stage. This stage is introduced only to transform frozen VFM representations into targets that are better matched to the compact tokenizer bottleneck, and is not part of the tokenizer training itself. In particular, its parameters are trained independently and are not shared with the final tokenizer. The motivation is twofold. First, raw VFM features are typically high-dimensional and channel-redundant, making them suboptimal as direct semantic supervision for a low-dimensional tokenizer bottleneck. Second, raw VFM features are often spatially imperfect at tokenizer resolution, which weakens their usefulness as structural targets. We therefore refine the frozen VFM into two fixed priors before tokenizer training: a compact semantic target for SCR and a spatially cleaner structural target for SSR.

This stage is applied to all considered frozen representation backbones, including MAE-L, SigLIP2-SO400M, DINOv3-L, and DINOv2-L, using the same training recipe unless otherwise specified. Given an input image \(x\), a frozen representation encoder produces a raw feature
\(
\mathbf{H}_{\mathrm{vfm}} \in \mathbb{R}^{N \times D}.
\)
We then learn a lightweight prior projector \(\mathcal{P}_{\theta}^{t}\) that maps the raw VFM feature into a compact bottleneck representation
\begin{equation}
\mathbf{Z}_T = \mathcal{P}_{\theta}^{t}(\mathbf{H}_{\mathrm{vfm}}) \in \mathbb{R}^{N \times d},
\end{equation}
where \(d=32\) for all backbones. To preserve the semantic content of the original representation under this compact bottleneck, a lightweight reconstruction decoder \(\mathcal{Q}_{\theta}^{t}\), implemented as a 4-layer ViT with hidden dimension 1024, reconstructs the raw representation as
\begin{equation}
\hat{\mathbf{H}}_{\mathrm{vfm}} = \mathcal{Q}_{\theta}^{t}(\mathbf{Z}_T).
\end{equation}
After training, the compact feature \(\mathbf{Z}_T\) and its globally pooled summary \(\mathbf{z}_{T,g}\) are used as the fixed semantic targets for SCR.

\begin{figure}[t]
    \centering
    \includegraphics[width=\textwidth]{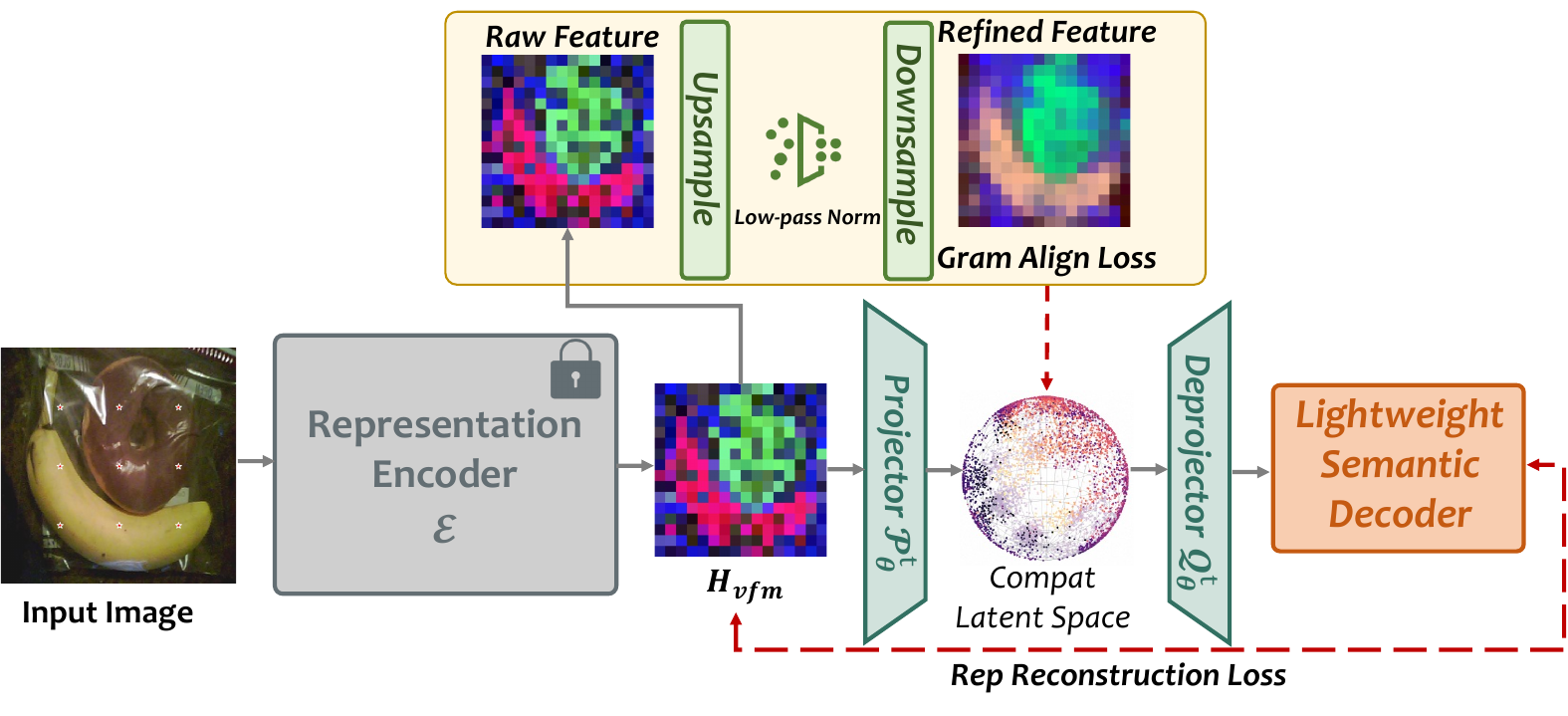}
    \caption{\textbf{Detailed pipeline of the VFM refine stage.}
Given an input image, a frozen representation encoder produces raw VFM features $H_{\mathrm{vfm}}$. A lightweight projector--deprojector pair compresses these features into a compact latent space and reconstructs them with a representation reconstruction loss, producing a bottleneck-matched semantic prior. To improve spatial suitability, the raw feature is also upsampled, low-pass normalized, and downsampled, and the compact feature is aligned to this refined structural target via a Gram loss. The refined semantic and structural priors are then fixed and used to supervise PAE training.}
    \label{fig:vfm_refine_pipe}
\end{figure}

\begin{figure}[t]
    \centering
    \includegraphics[width=\textwidth]{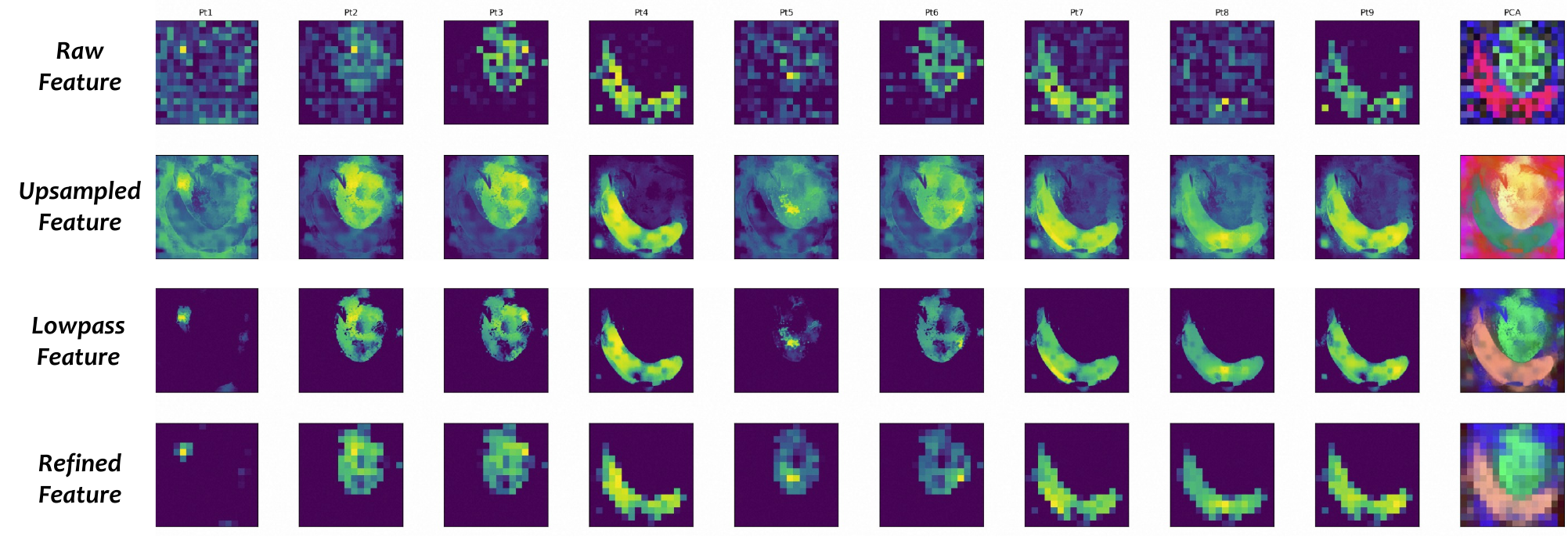}
    \caption{\textbf{Visualization of the structural refinement process.}
From left to right, we show the raw VFM feature, the AnyUp-upsampled feature, the low-pass normalized feature, and the final refined feature after resizing back to tokenizer resolution. The refinement suppresses noisy local variation while preserving coarse spatial structure, yielding a cleaner target for SSR.}
    \label{fig:vis_refine}
\end{figure}

In parallel, we construct a refined structural reference directly from the raw VFM feature. Specifically, \(\mathbf{H}_{\mathrm{vfm}}\) is first upsampled using a pretrained AnyUp~\cite{anyup} model to obtain a dense spatial feature map, then processed with the low-pass spatial normalization procedure of~\cite{irepa}, and finally resized back to the tokenizer resolution using bilinear interpolation. We denote the resulting refined structural feature by \(\mathbf{H}_{\mathrm{ref}}\). This spatial refinement suppresses noisy local variation while preserving coarse patch-wise spatial relations, yielding a cleaner structural reference for subsequent structure alignment.

The overall objective of the refine stage is
\begin{equation}
\mathcal{L}_{\mathrm{refine}}
=
\lambda_{\mathrm{rep}} \mathcal{L}_{\mathrm{rep}}
+
\lambda_{\mathrm{gram}} \mathcal{L}_{\mathrm{gram}},
\end{equation}
where \(\lambda_{\mathrm{rep}}=\lambda_{\mathrm{gram}}=1.0\) in all experiments.

The first term is a representation reconstruction loss:
\begin{equation}
\mathcal{L}_{\mathrm{rep}}
=
\left\|
\hat{\mathbf{H}}_{\mathrm{vfm}}-\mathbf{H}_{\mathrm{vfm}}
\right\|_2^2.
\end{equation}
Its role is to preserve the native semantic content of the frozen VFM while forcing the representation through a compact tokenizer-compatible bottleneck. In this way, \(\mathbf{Z}_T\) becomes a bottleneck-matched semantic prior rather than a direct copy of the original high-dimensional feature.

The second term improves the spatial suitability of the compact target through Gram-based structure alignment:
\begin{equation}
\mathcal{L}_{\mathrm{gram}}
=
\left\|
\mathrm{Gram}(\mathbf{Z}_T) - \mathrm{Gram}(\mathbf{H}_{\mathrm{ref}})
\right\|_F^2.
\end{equation}
Here \(\mathrm{Gram}(\cdot)\) denotes the patch-wise Gram matrix computed from channel-normalized token features, so the compared features need not share the same channel dimension. This objective encourages the compact target \(\mathbf{Z}_T\) to preserve the coarse spatial relations of the refined structural reference \(\mathbf{H}_{\mathrm{ref}}\), making it more suitable for subsequent structure-aware supervision. During tokenizer training, SSR uses the fixed Gram statistics of \(\mathbf{H}_{\mathrm{ref}}\) as its structural target.

Figure~\ref{fig:vfm_refine_pipe} illustrates the detailed pipeline of this refine stage. Starting from the frozen VFM feature, the prior projector--decoder pair learns a compact bottleneck representation aligned with the original representation content, while the spatial refinement branch constructs a cleaner structural reference through upsampling, low-pass normalization, and downsampling. Together, these two paths produce the fixed semantic and structural priors used in the final PAE training.

Unless otherwise specified, the refine stage is trained for 16 epochs using AdamW with \(\beta_1=0.9\) and \(\beta_2=0.98\), a global batch size of 1024, EMA decay 0.9978, and a cosine learning-rate schedule from \(2\times10^{-4}\) to \(2\times10^{-5}\) with 1 warmup epoch. We use AnyUp as the feature upsampler, set the low-pass strength to 0.4 for all backbones, and use bilinear interpolation for the final downsampling step. The input image resolution follows the underlying frozen encoder, as summarized in Table~\ref{tab:vfm_refine_config}. After training, all refined targets are fixed during subsequent tokenizer training: \(\mathbf{Z}_T\) and \(\mathbf{z}_{T,g}\) provide semantic supervision for SCR, while the Gram statistics derived from \(\mathbf{H}_{\mathrm{ref}}\) provide the structural supervision for SSR.

\begin{table}[t]
\centering
\caption{\textbf{Configurations for the VFM refine stage.}}
\renewcommand{\arraystretch}{1.15}
\setlength{\tabcolsep}{6pt}
\begin{tabular}{l|cccc}
\toprule
encoder & MAE-L & SigLIP2-SO400M & DINOv3-L & DINOv2-L \\
\midrule
input image size & 256 & 256 & 256 & 224 \\
latent dimension & \multicolumn{4}{c}{32} \\
decoder depth & \multicolumn{4}{c}{4} \\
decoder hidden dim & \multicolumn{4}{c}{1024} \\
training epochs & \multicolumn{4}{c}{16} \\
EMA decay & \multicolumn{4}{c}{0.9978} \\
global batch size & \multicolumn{4}{c}{1024} \\
optimizer & \multicolumn{4}{c}{AdamW, $\beta_1{=}0.9$, $\beta_2{=}0.98$} \\
learning rate & \multicolumn{4}{c}{$2\times10^{-4}$ cosine decay to $2\times10^{-5}$} \\
warmup epochs & \multicolumn{4}{c}{1} \\
active losses & \multicolumn{4}{c}{$\lambda_{\mathrm{rep}}{=}1.0$, $\lambda_{\mathrm{gram}}{=}1.0$} \\
upsampler & \multicolumn{4}{c}{AnyUp} \\
upsample size & \multicolumn{4}{c}{256} \\
low-pass strength & \multicolumn{4}{c}{0.4} \\
downsample type & \multicolumn{4}{c}{bilinear} \\
\bottomrule
\end{tabular}
\label{tab:vfm_refine_config}
\end{table}

\subsection{Sampling and Evaluation Protocol}
We follow the evaluation protocol commonly used in prior representation-native AE works~\cite{fae, gae}. For sampling without classifier-free guidance (CFG), we use an SDE-based sampler; when CFG is enabled, we instead use an ODE-based sampler. Unless otherwise specified, all reported generative results are obtained with 250 sampling steps.
For ImageNet evaluation, we use class-uniform sampling so that each category contributes the same number of generated samples, consistent with prior work~\cite{fae,gae,rae}. For models with latent dimension $d=32$, the sampling hyperparameters depend on the training duration. At 800 epochs, we use a time shift of 0.4, a CFG interval of 0.3, and a guidance scale of 3.3. For the 80-epoch checkpoints, we use a time shift of 0.4, a CFG interval of 0.25, and a guidance scale of 2.5. Unless otherwise specified, both gFID and rFID are computed using 50,000 images. Reconstruction quality is measured on reconstructed validation images, whereas generation quality is measured on synthesized samples obtained under the corresponding sampling setup.

\subsection{Detailed Configuration of Ablation Study}
\subsubsection{Pilot Studies}
\label{app:pilot_studies}

The pilot studies in \fref{fig:pilot} are designed as controlled experiments to examine how different latent-manifold properties relate to downstream generation quality. Each group isolates one factor while keeping the tokenizer scaffold, optimization setup, downstream generator, and evaluation protocol fixed within the group. Specifically, Group~1 varies the bottleneck channel dimension to study the mismatch between reconstruction fidelity and generation quality; Group~2 varies only the SSR weight to probe spatial structure; Group~3 varies only the MCR weight to probe local continuity; and Group~4 varies only the SCR weight to probe global semantic organization. To avoid confounding semantic supervision with patch-level structural alignment, the SCR objective in Group~4 is applied only to the globally pooled token rather than to patch tokens.

Unless otherwise specified, all groups use the same ViT-AE-Large tokenizer scaffold on ImageNet $256{\times}256$, with latent resolution $16{\times}16$. All tokenizers are trained for 15 epochs with a global batch size of 256 using AdamW with $\beta_1{=}0.9$ and $\beta_2{=}0.98$, and a cosine learning-rate schedule from $2{\times}10^{-4}$ to $2{\times}10^{-5}$. All pilot-study tokenizers use the same reconstruction objective consisting only of L1 and LPIPS losses, without GAN loss. All downstream evaluations are conducted with the same LightningDiT-XL/1 generator setup within each group, and generation quality is measured by gFID on 10K generated samples. In addition, all tokenizer variants are evaluated using the proposed latent diagnostics SSC, LPC, and GSQ.
\paragraph{Group 1: Reconstruction vs.\ generation (rFID vs.\ gFID).}
The first group studies whether stronger reconstruction alone necessarily leads to better generation. We train a plain ViT-AE using only L1 and LPIPS losses, without any prior-alignment or manifold regularization terms. The only factor varied in this group is the bottleneck channel dimension, with $d \in \{32,48,64,96,128\}$.
All other settings, including tokenizer architecture, latent resolution, optimizer, training schedule, and downstream LightningDiT-XL/1 generator, are kept fixed. This group is designed to change reconstruction capacity while minimizing changes to the rest of the training pipeline. As the bottleneck dimension increases, reconstruction quality improves monotonically, but generation quality does not necessarily improve in the same way, revealing a mismatch between reconstruction fidelity and downstream learnability, as shown in \fref{fig:pilot}(a).
\paragraph{Group 2: Spatial structure (SSC vs.\ gFID).}
The second group isolates the effect of spatial structure. Starting from the same baseline tokenizer with latent shape $16{\times}16{\times}32$ and L1+LPIPS reconstruction losses, we activate only the SSR term and sweep its weight over $\lambda_{\mathrm{SSR}} \in \{0, 0.05, 0.1, 0.2, 0.5\}$. To maintain a single-factor intervention, we fix $\lambda_{\mathrm{MCR}}{=}0$ and disable SCR throughout this group. Under this design, the primary change is in structure-aware supervision, while local continuity and semantic supervision remain absent. We therefore use this group to probe how spatial token organization, as measured by SSC, relates to downstream generation quality (\fref{fig:pilot}(b)).
\paragraph{Group 3: Local continuity (LPC vs.\ gFID).}
The third group focuses on local continuity of the latent manifold. Using the same baseline tokenizer as in Group~2, we activate only the MCR term and sweep its weight over $\lambda_{\mathrm{MCR}} \in \{0, 0.05, 0.15, 0.3, 0.5\}$, while fixing $\lambda_{\mathrm{SSR}}{=}0$ and disabling SCR. Under this controlled setting, the main intervention is local perturbation regularization, without additional spatial-structure or semantic-alignment supervision. This allows us to examine how changes in local manifold smoothness, as measured by LPC, affect downstream generation quality (\fref{fig:pilot}(c)).
\paragraph{Group 4: Global semantics (GSQ vs.\ gFID).}
The fourth group isolates the role of global semantic organization. We again use the same baseline tokenizer as in Group~2, but now activate only the SCR term and sweep its weight over $\lambda_{\mathrm{SCR}} \in \{0, 0.1, 0.3, 0.6, 1.0\}$, while fixing $\lambda_{\mathrm{SSR}}{=}0$ and $\lambda_{\mathrm{MCR}}{=}0$. To avoid confounding semantic supervision with patch-level structural alignment, SCR is applied only to the globally pooled token in this group, rather than to patch tokens. As a result, this intervention primarily changes global semantic supervision while minimizing its effect on spatial token layout. We therefore use this group to probe how global semantic organization, as measured by GSQ, relates to downstream generation quality (\fref{fig:pilot}(d)).

\paragraph{Ablation on prior alignment.}
This ablation is conducted on PAE (DINOv2). We follow the same tokenizer training setting as in the main experiments, except that the tokenizer is trained for 25 epochs instead of 50. The downstream class-conditional generation setup is kept identical to the main experiments, including the same LightningDiT-XL/1 training and evaluation protocol. This ensures that the comparison isolates the effect of prior-target construction rather than differences in downstream generator optimization.

\subsubsection{Ablation on regularization strategy}
\label{app:core_ablation_settings}

Table~\ref{tab:ablation_core}(a) compares our prior-alignment design against several generic latent regularization baselines under the same tokenizer scaffold and downstream generator setting. Unless otherwise specified, all variants are built on top of the same PAE backbone and share the same reconstruction objective, optimizer, training schedule, latent shape, and LightningDiT-XL/1 evaluation protocol as the main experiment.

\paragraph{Baseline 1: PAE without $\loss_p$.}
This baseline removes the full prior-alignment loss $\loss_p$, i.e., no SSR, MCR, or SCR is applied during tokenizer training. It serves as the reference model for measuring the contribution of prior alignment.

\paragraph{Baseline 2: PAE without $\loss_p$ + KL regularization.}
Starting from Baseline 1, we add a weak KL penalty on the latent representation:
\begin{equation}
\mathcal{L}_{\mathrm{KL}}
=
D_{\mathrm{KL}}\!\left(q(\mathbf{z}\mid \mathbf{x}) \,\|\, \mathcal{N}(0,I)\right),
\end{equation}
with weight $10^{-6}$. This baseline is intended to test whether a generic distributional regularizer that mildly constrains the latent space can improve downstream generation without using geometry-targeted prior alignment.

\paragraph{Baseline 3: PAE without $\loss_p$ + diffusion-loss regularization.}
Starting from Baseline 1, we attach a lightweight diffusion regularizer branch on top of the latent tokens, implemented as a 2-layer DiT block. This branch is trained to predict the standard diffusion target on noisy latent codes, following the spirit of Unified Latent~\cite{ul}. The resulting auxiliary diffusion loss is added only during tokenizer training and is not used at inference time. This baseline tests whether directly encouraging diffusion compatibility through an auxiliary latent diffusion objective can replace our explicit manifold-oriented prior alignment.

In all three cases, the downstream latent diffusion generator is trained from scratch using the same protocol as in the main experiment. This ensures that the comparison isolates the effect of tokenizer-side regularization rather than differences in generator capacity or training setup.

\paragraph{Ablation on other design choices.}
These ablations are also performed on PAE (DINOv2). Unless otherwise specified, all tokenizer, generator, sampling, and evaluation settings are exactly the same as those used in the main experiments, with only the ablated design choice changed.


\section{More Ablation and Discussion}
\label{app:ablation}

\subsection{Pearson correlation analysis for Manifold Metrics}
\label{app:cross_tokenizer_metric_validation}

\begin{figure}[ht]
    \centering
    \includegraphics[width=\textwidth]{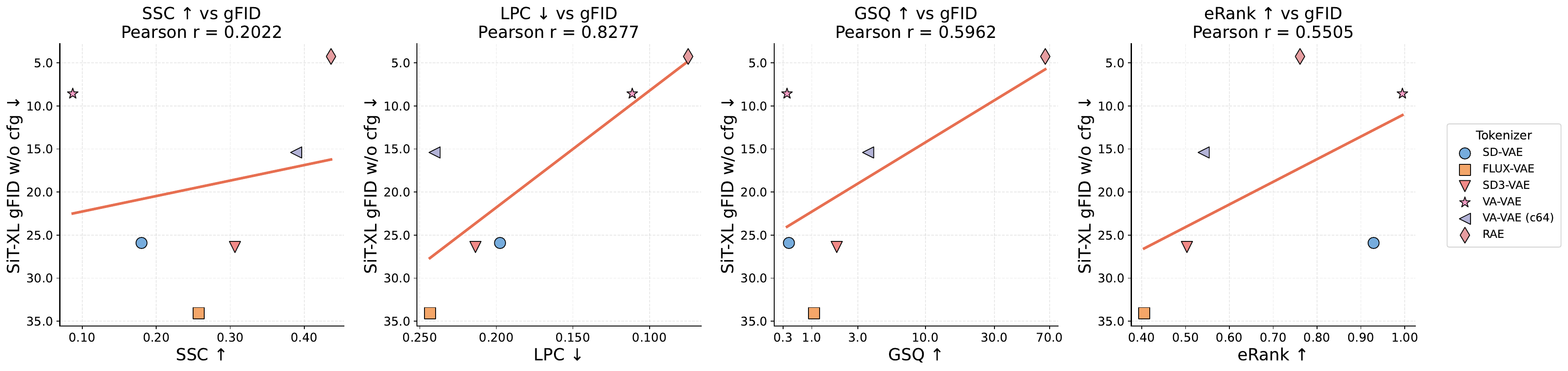}
    \caption{\textbf{Cross-tokenizer validation of manifold metrics.}
    Relationship between the proposed latent-space metrics and downstream SiT-XL~\cite{ma2024sit} generation quality (without classifier-free guidance) across a diverse set of existing autoencoders and tokenizers. To make the trend direction easy to interpret, the correlations are computed after simple coordinate normalization, so that a positive slope or Pearson coefficient consistently means that a better metric value aligns with better generation quality. Although tokenizer families differ substantially in architecture and latent parameterization, the overall trends remain directionally consistent, supporting that these metrics capture general properties relevant to diffusion performance rather than artifacts specific to our own setting.}
    \label{fig:person_fig_gFID_sit_xl}
\end{figure}

To further verify that the proposed manifold metrics are not specific to our own tokenizer design, we evaluate their relationship with downstream generation quality across a diverse set of existing autoencoders and tokenizers, including VAE-based models, vector-quantized tokenizers, masked-token methods, and representation autoencoders. Figure~\ref{fig:person_fig_gFID_sit_xl} shows the relationship between each metric and the resulting SiT-XL gFID without classifier-free guidance.

For ease of interpretation, we normalize the plot coordinates before computing the regression line and Pearson correlation, so that the sign of the trend is consistent across metrics. Under this convention, a positive slope or Pearson coefficient can always be read as: better latent geometry is associated with better downstream generation quality. Concretely, this normalization uses sign flips for gFID and LPC, and a log transform for GSQ to reduce scale skew, while the displayed axes are formatted in their usual readable forms.

With this convention in mind, the observed trends are directionally consistent with our design motivation. Metrics that characterize more diffusion-friendly latent geometry tend to align with better generative performance across different tokenizer families, rather than only within our own method. Among the four metrics, LPC exhibits the clearest monotonic relation, suggesting that local path continuity is strongly associated with diffusion quality across tokenizers. This is consistent with the intuition that smoother local transitions in latent space make denoising trajectories easier to model. GSQ and eRank also show meaningful positive correlations, indicating that semantic neighborhood quality and effective latent utilization are both relevant to downstream generation. SSC displays a weaker correlation in this cross-tokenizer comparison, which suggests that global structural consistency alone may be insufficient to explain generation quality when tokenizer families differ substantially in architecture, compression ratio, and latent dimensionality.

Importantly, these results provide evidence beyond the setting used in our main experiments. The metrics are computed independently on each tokenizer, while the generation quality is measured after training diffusion transformers on the corresponding latent spaces. Therefore, the observed correlations indicate that these manifold properties are not merely artifacts of our particular tokenizer or training recipe, but reflect broader characteristics that influence diffusion performance across heterogeneous latent representations.

At the same time, we do not claim that any single metric fully determines generation quality. The absolute strength of correlation can vary across tokenizer families, and some metrics may also be affected by factors such as latent dimensionality, representation scale, and the inductive bias of the encoder-decoder architecture. For this reason, we view these metrics as complementary descriptors of diffusion-friendliness rather than isolated predictors. Taken together, the cross-tokenizer evidence supports the usefulness and generality of the proposed analysis framework.

\subsection{Few-Step Sampling Results}

We further evaluate few-step sampling under the same setting as the long-training comparison in Table~\ref{tab:few_step_sampling}, using \pae{} (DINOv2) with LightningDiT-XL/1 trained for 800 epochs and classifier-free guidance. For a fair comparison, we compare against FAE under the same generator setting.

\begin{figure}[ht]
    \centering
    \includegraphics[width=\textwidth]{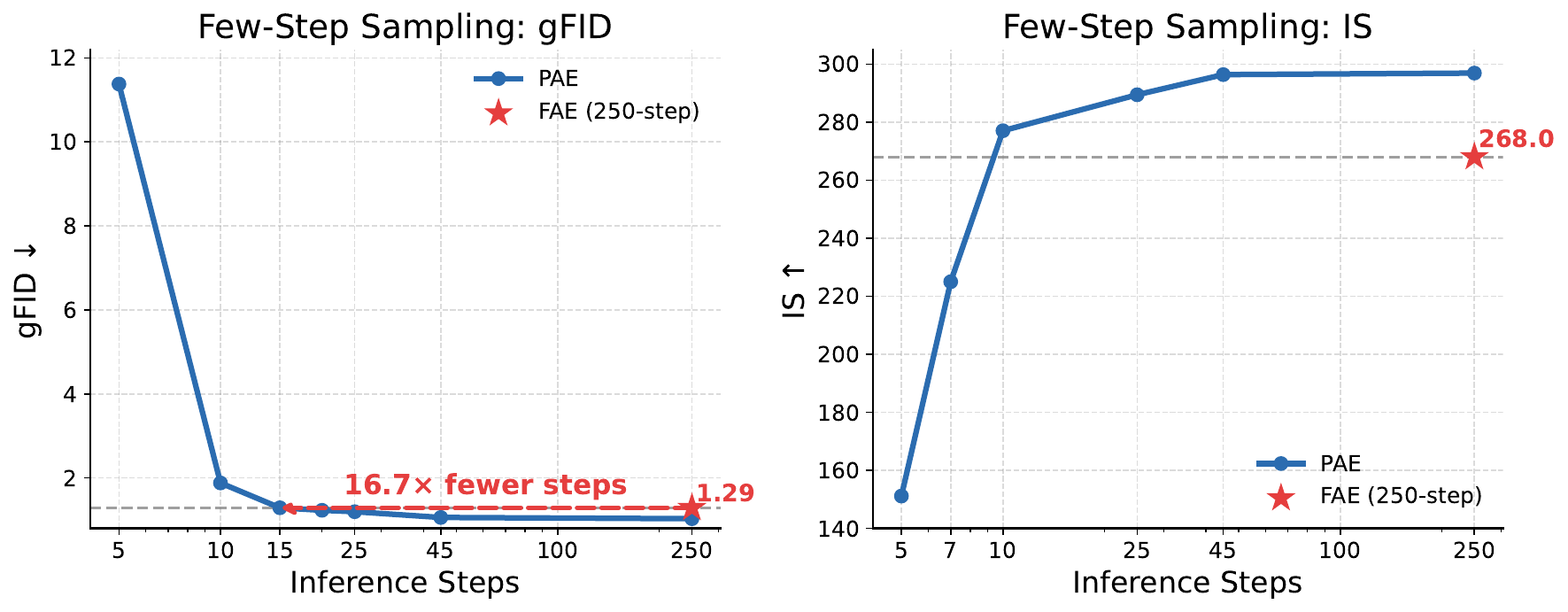}
    \caption{\textbf{Few-step sampling performance.}
    Results are reported for \pae{} (DINOv2) with LightningDiT-XL/1 trained for 800 epochs using classifier-free guidance; FAE is evaluated under the same setting. Left: gFID versus inference steps. Right: IS versus inference steps. \pae{} quickly approaches its full-sampling performance and matches the 250-step gFID of FAE using only 15 steps, corresponding to 16.7$\times$ fewer inference steps. It also achieves consistently higher IS than FAE in the few-step regime.}
    \label{fig:few_step_sampling}
\end{figure}

Figure~\ref{fig:few_step_sampling} and Table~\ref{tab:few_step_sampling} show that \pae{} quickly approaches its full-sampling performance as the number of inference steps increases. In particular, \pae{} matches the 250-step gFID of FAE in only 15 steps, corresponding to 16.7$\times$ fewer inference steps. Moreover, \pae{} achieves a gFID of 1.05 at 45 sampling steps, which is the few-step result reported in the introduction. \pae{} also achieves substantially higher IS than FAE in the few-step regime, indicating that the learned latent space is favorable for efficient diffusion sampling.

\begin{table}[t]
\centering
\caption{\textbf{Few-step sampling under the same long-training setting.}
Results are reported for \pae{} (DINOv2) with LightningDiT-XL/1 trained for 800 epochs using classifier-free guidance. FAE is evaluated under the same generator setting.}
\label{tab:few_step_sampling}
\small
\setlength{\tabcolsep}{6pt}
\renewcommand{\arraystretch}{1.05}
\begin{tabular}{lccc}
\toprule
\textbf{Method} & \textbf{Steps} & \textbf{gFID}$\downarrow$ & \textbf{IS}$\uparrow$ \\
\midrule
\pae{} (DINOv2) & 10  & 1.88 & 277.0 \\
\pae{} (DINOv2) & 15  & 1.28 & 282.0 \\
\pae{} (DINOv2) & 25  & 1.20 & 289.4 \\
\pae{} (DINOv2) & 45  & 1.06 & 296.4 \\
\pae{} (DINOv2) & 250 & 1.03 & 296.9 \\
\midrule
FAE             & 250 & 1.29 & 268.0 \\
\bottomrule
\end{tabular}
\end{table}

\subsection{Why does \pae{} achieve a better fidelity--learnability balance?}

\paragraph{Breaking the fidelity--learnability trade-off.}
Figure~\ref{fig:metrics_comp}\textnormal{(a)} compares tokenizers in terms of reconstruction fidelity and downstream learnability. Existing methods exhibit a clear trade-off. Some tokenizers favor reconstruction but remain harder for diffusion to learn, while others improve learnability at the cost of weaker reconstruction. In contrast, \pae{} achieves strong reconstruction together with the best learnability, indicating that its gain does not come from sacrificing one side of the trade-off.

\paragraph{Balanced latent geometry rather than a single dominant factor.}
Figure~\ref{fig:metrics_comp}\textnormal{(b)} helps explain this advantage. Compared with prior tokenizers, \pae{} is simultaneously strong on the three geometry dimensions most relevant to diffusion, namely spatial structure (SSC), local continuity (LPC), and global semantics (GSQ), while also maintaining high latent utilization as measured by eRank. This suggests that \pae{} succeeds not because of a single dominant property, but because it constructs a more balanced latent manifold that is structurally coherent, locally smooth, and semantically organized. Notably, some tokenizers such as RAE remain competitive on several geometry dimensions. However, their much higher-dimensional latent space leads to weaker effective utilization under a fixed generator budget, which helps explain why their downstream generation still falls behind more compact tokenizers such as \pae{}.

\paragraph{Why DINO $>$ SigLIP $>$ MAE?}
Figure~\ref{fig:metrics_comp}\textnormal{(c)} further shows that different VFM backbones induce different geometry profiles even under the same tokenizer design. DINO-based \pae{} is the most balanced across SSC, LPC, and GSQ, which is consistent with its strongest downstream generation. SigLIP-based \pae{} achieves the strongest semantic organization but weaker spatial structure and only moderate continuity, which explains why it remains competitive but does not match DINO-based \pae{}. MAE-based \pae{} retains reasonable spatial structure but is clearly weaker in continuity and semantics, which aligns with its worse generation quality. Since eRank remains relatively close across these encoders under the same channel budget, the performance gap is better explained by differences in the three primary geometry properties than by latent utilization alone.

\subsection{Ablation on perturbation design for MCR}
\label{app:mcr_ablation}

We compare different perturbation designs under the same tokenizer backbone and training setup to test whether MCR helps through explicit continuity regularization rather than generic robustness.
All variants use the same reconstruction, SSR, and SCR objectives, and differ only in the perturbation design.
Perturbations are applied in the RMS-normalized sphere-like latent space along a random normalized direction.
For Small Perturb, the maximum angular deviation is $42.5^\circ$; for Large Perturb, it is $85^\circ$; and Cascaded Perturb (ours) uses both levels with the progressive consistency objective in Eq.~(4). Table~\ref{tab:mcr_design} shows that generic perturbation consistency already improves over removing MCR, confirming that local latent regularization is useful.
However, the perturbation design is important: small perturbations improve LPC and gFID but are limited in effect, while large perturbations slightly improve LPC at the cost of worse reconstruction.
In contrast, the proposed cascaded design achieves the best LPC and gFID without sacrificing rFID, indicating that MCR works by progressively regularizing local neighborhoods rather than by simply adding generic perturbation robustness.

\begin{table}[ht]
\centering
\caption{\textbf{Ablation on perturbation design for MCR.}
Generic perturbation consistency helps, but the proposed cascaded design gives the best continuity and generation quality.}
\label{tab:mcr_design}
\small
\setlength{\tabcolsep}{5pt}
\renewcommand{\arraystretch}{1.08}
\begin{tabular}{lccccc}
\toprule
\textbf{Method} & \textbf{Perturbation} & \textbf{LPC}$\downarrow$ & \textbf{rFID}$\downarrow$ & \textbf{gFID}$\downarrow$ & \textbf{IS}$\uparrow$ \\
\midrule
No Perturb                  & $0^\circ$                 & 0.258 & 0.25 & 2.10 & 188.9 \\
Small Perturb               & $42.5^\circ$        & 0.219 & 0.26 & 2.00 & 193.4 \\
Large Perturb               & $85^\circ$          & 0.205 & 0.28 & 2.04 & 191.6 \\
Cascaded Perturb (ours)     & $42.5^\circ+85^\circ$ & \textbf{0.170} & 0.26 & \textbf{1.80} & \textbf{218.3} \\
\bottomrule
\end{tabular}
\end{table}

\subsection{Cross-Encoder Quantitative Results.}
Table~\ref{tab:encoder_generalization} shows that the benefit of prior alignment is consistent across different frozen teachers.
Adding $\mathcal{L}_{\mathrm{p}}$ substantially improves both gFID and IS for all four encoders, confirming that the proposed manifold-shaping objectives are not tied to a particular representation backbone.
Among them, DINOv2 and DINOv3 achieve the strongest overall performance, while SigLIP2 remains competitive and MAE also benefits noticeably despite a weaker starting point.
These results suggest that PAE generalizes well across diverse pretrained feature spaces, while stronger teacher representations still lead to better final generative quality.

\begin{table}[ht]
\centering
\caption{\textbf{Encoder generalization across frozen teachers.} 
PAE consistently improves over the corresponding tokenizer scaffold without $\mathcal{L}_{\mathrm{p}}$ across DINOv2, SigLIP2, DINOv3, and MAE.}
\label{tab:encoder_generalization}
\small
\setlength{\tabcolsep}{3pt}
\renewcommand{\arraystretch}{1.05}
\begin{tabular}{@{}cccccc@{}}
\toprule
\textbf{Metric} & \textbf{Setting} & \textbf{DINOv2} & \textbf{SigLIP2} & \textbf{DINOv3} & \textbf{MAE} \\
\midrule
\multirow{2}{*}{gFID$\downarrow$}
& w/o $L_{\mathrm{p}}$ & 7.79 & 6.89 & 6.62 & 7.97 \\
& w/ $L_{\mathrm{p}}$  & 1.80 & 2.32 & 1.81 & 3.65 \\
\midrule
\multirow{2}{*}{IS$\uparrow$}
& w/o $L_{\mathrm{p}}$ & 117.2 & 123.69 & 124.37 & 116.93 \\
& w/ $L_{\mathrm{p}}$  & 218.3 & 199.6 & 216.72 & 156.9 \\
\bottomrule
\end{tabular}
\end{table}



\section{More Visualizations}
\label{app:qualitative}

\subsection{Additional Reconstruction Visualizations}
\label{app:recon_vis}
\begin{figure}[t]
    \centering
    \includegraphics[width=\textwidth]{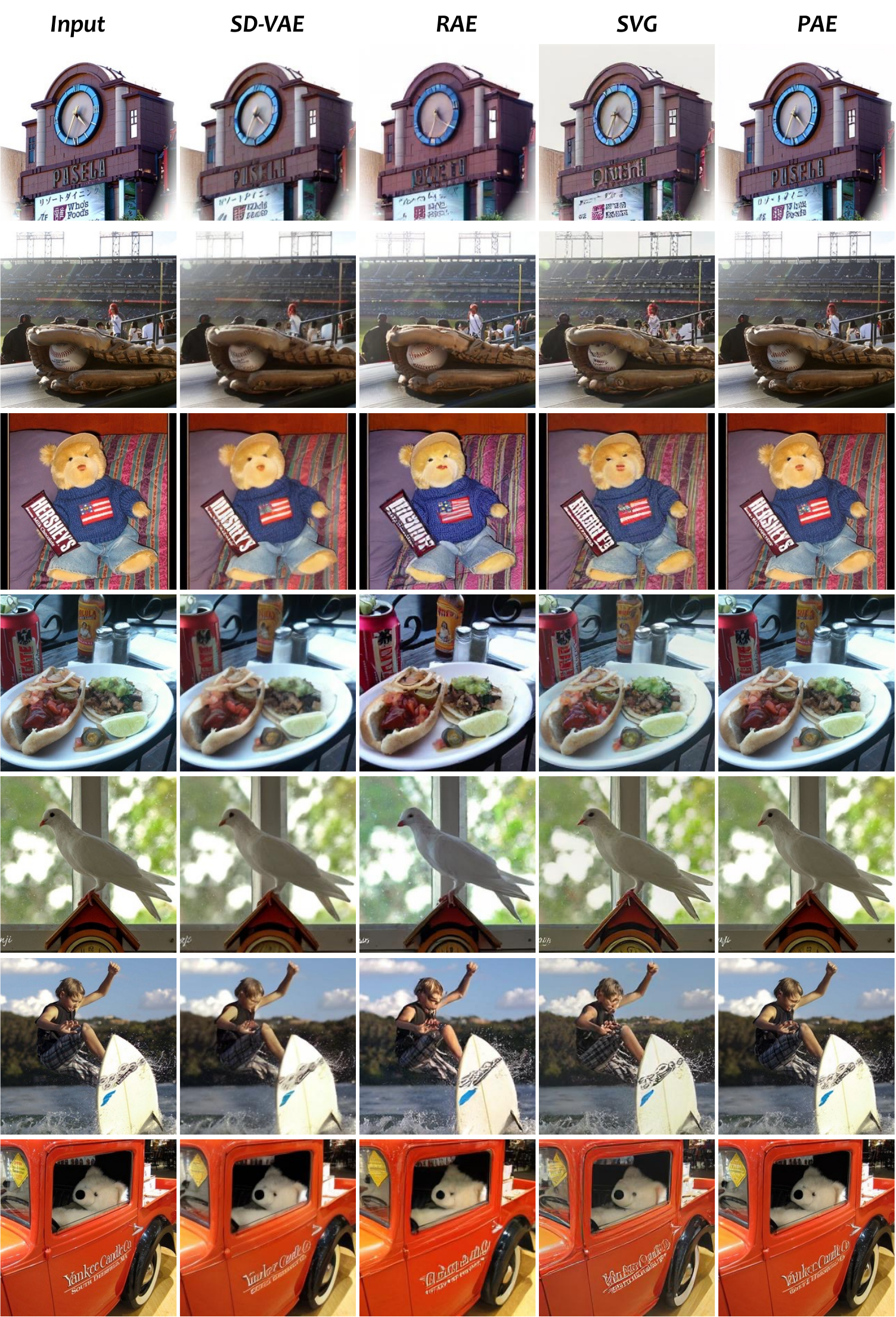}
    \caption{\textbf{Additional reconstruction comparisons.}
PAE consistently preserves finer visual details than representative tokenizer baselines.}
    \label{fig:recon_vis1}
\end{figure}

\begin{figure}[t]
    \centering
    \includegraphics[width=\textwidth]{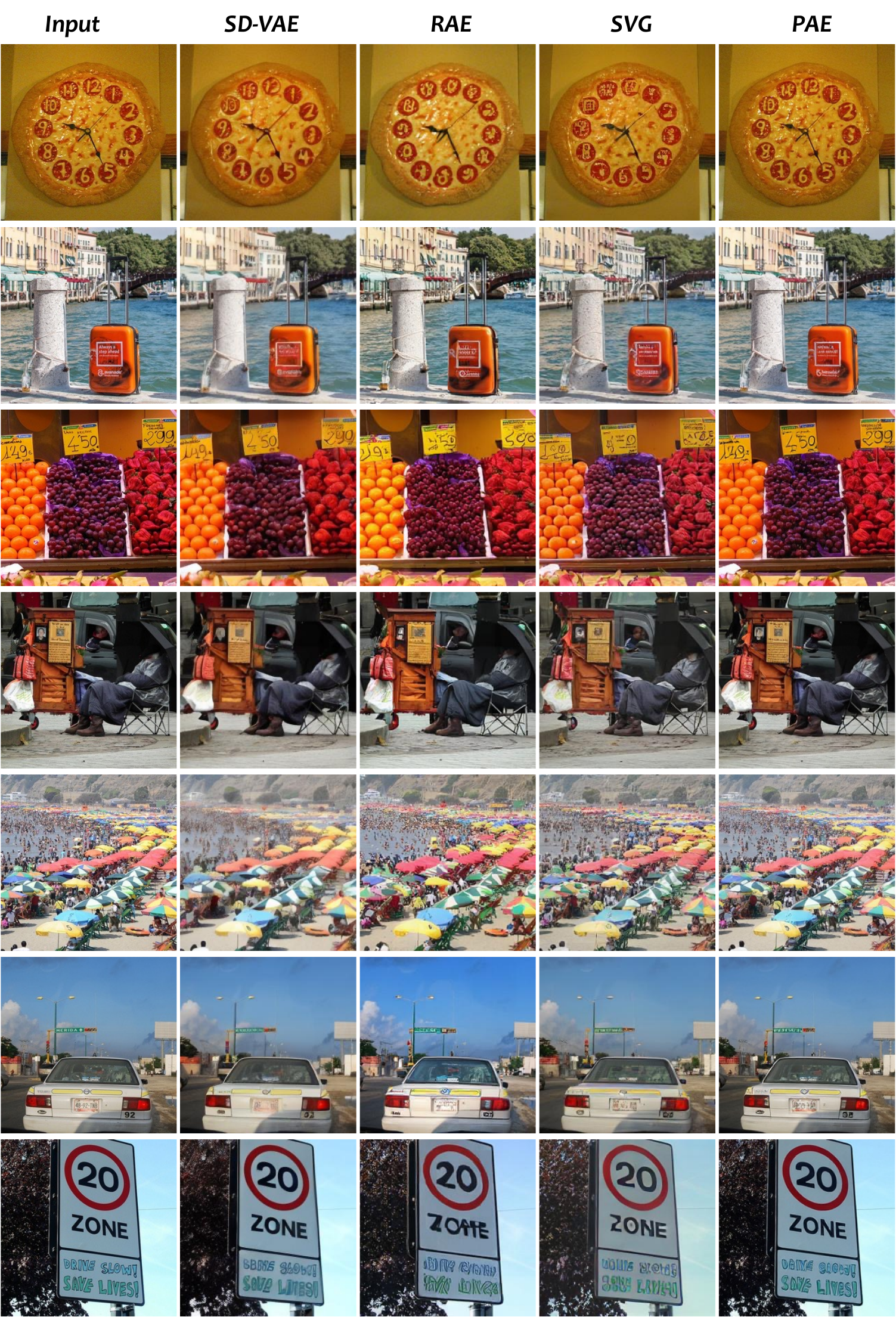}
    \caption{\textbf{Additional reconstruction comparisons.}
PAE consistently preserves finer visual details than representative tokenizer baselines.}
    \label{fig:recon_vis2}
\end{figure}

Figures~\ref{fig:recon_vis1} and \ref{fig:recon_vis2} provide additional qualitative comparisons of reconstruction performance across various tokenizer architectures. As shown in these results, PAE consistently achieves superior reconstruction fidelity compared to existing methods like SD-VAE and RAE. Notably, PAE excels at preserving fine-grained high-frequency details, such as thin structural lines and complex textual information, which are often blurred or lost in reconstruction-oriented baselines.

\subsection{Spatial Structure Visualizations}
\label{app:spatial_structure_vis}
\begin{figure}[t]
    \centering
    \includegraphics[width=\textwidth]{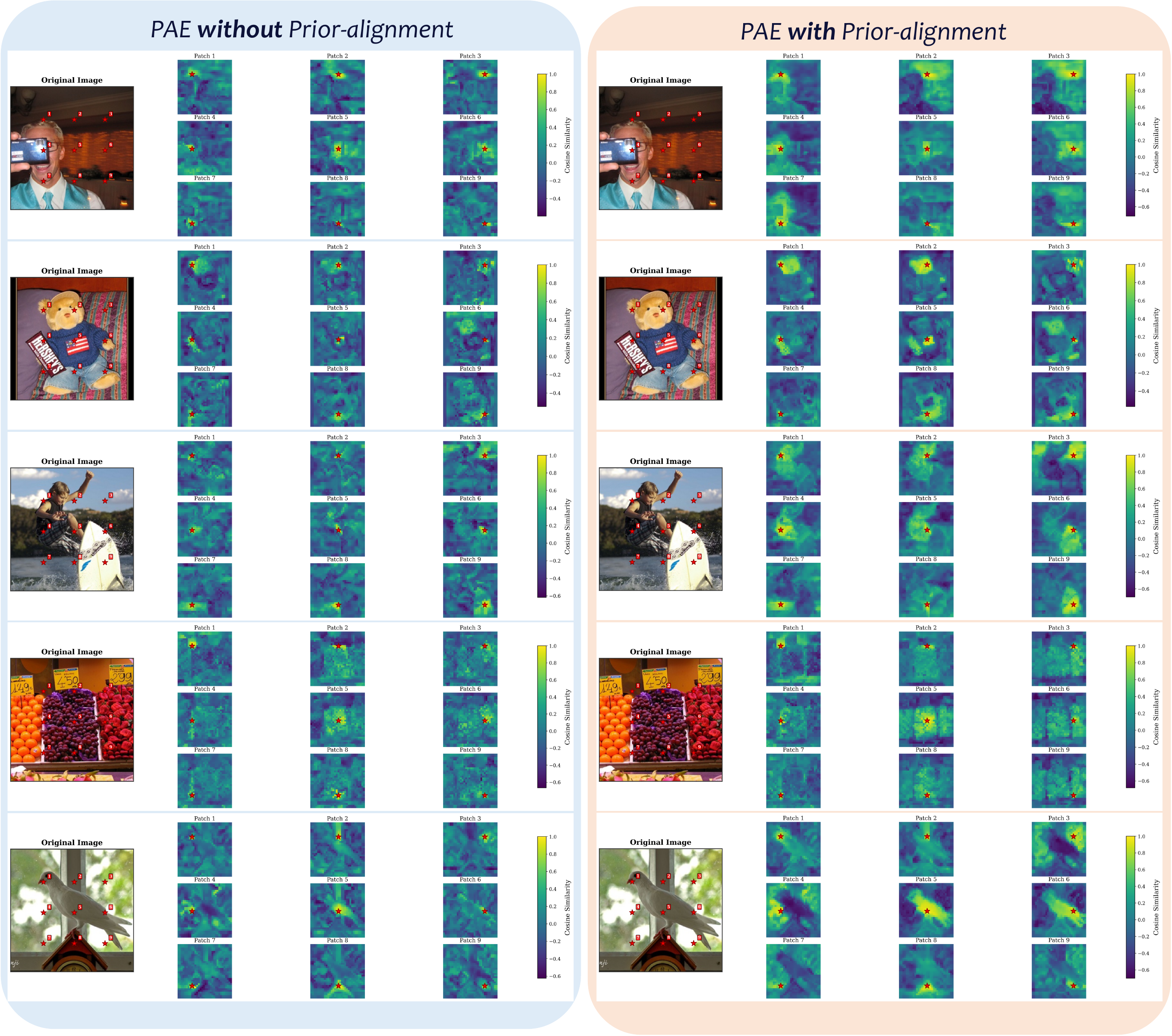}
    \caption{\textbf{Spatial structure visualization.}
    With prior alignment, PAE produces clearer patch-wise similarity structure and better preserves object-level spatial relations.}

    \label{fig:spatial_vis}
\end{figure}

Figure~\ref{fig:spatial_vis} visualizes the spatial structure preservation of PAE using patch-wise similarity maps. Through Spatial Structure Regularization (SSR), PAE aligns its latent spatial relations with refined VFM-derived structural priors. The visualization demonstrates that PAE captures much clearer instance-level boundaries and structural consistency compared to versions without prior-alignment, effectively improving the Spatial Structure Coherence (SSC) metric and enabling the diffusion model to focus on generative patterns rather than compensating for spatial misalignment.

\subsection{Global Semantic Organization}
\label{app:global_semantic_organization}
\begin{figure}[t]
    \centering
    \includegraphics[width=\textwidth]{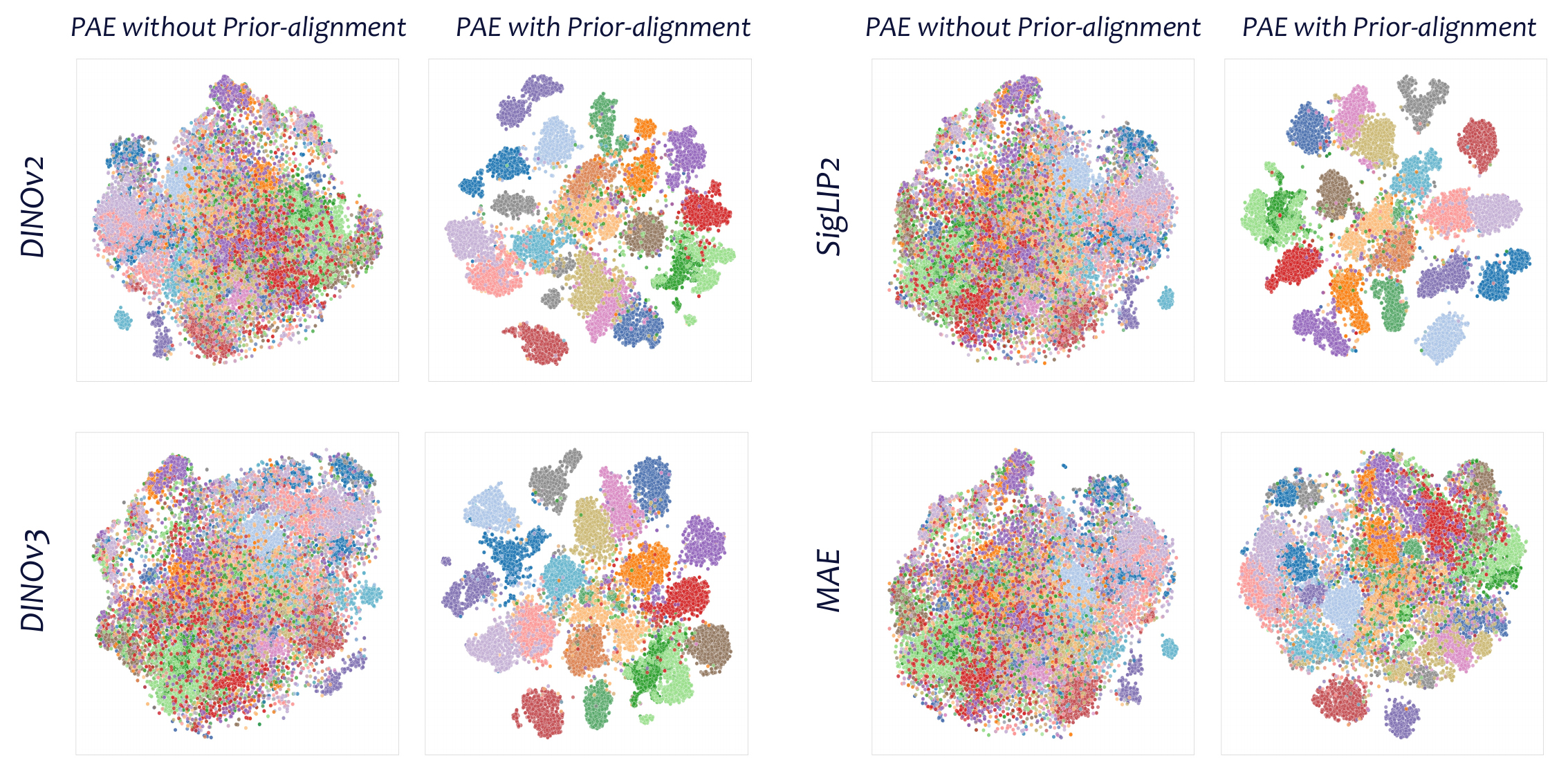}
    \caption{\textbf{Global semantic organization.}
    Prior alignment yields more compact and class-consistent latent neighborhoods, improving the semantic organization of the tokenizer manifold.}
    \label{fig:global_semantic_organization}
\end{figure}

The impact of Semantic Consistency Regularization (SCR) on the global organization of the latent manifold is visualized in Figure~\ref{fig:global_semantic_organization}. By aligning the tokenizer representation with compressed VFM semantic priors, PAE (with Prior-alignment) exhibits significantly tighter class-wise clustering in the latent space. This improved Global Semantic Quality (GSQ) simplifies conditional generative modeling by ensuring that semantically similar samples are compactly organized, facilitating faster convergence and better final generation quality as evidenced in our ImageNet experiments.

\subsection{DiT Latent Interpolation}
\label{app:dit_interp}
\begin{figure}[t]
    \centering
    \includegraphics[width=\textwidth]{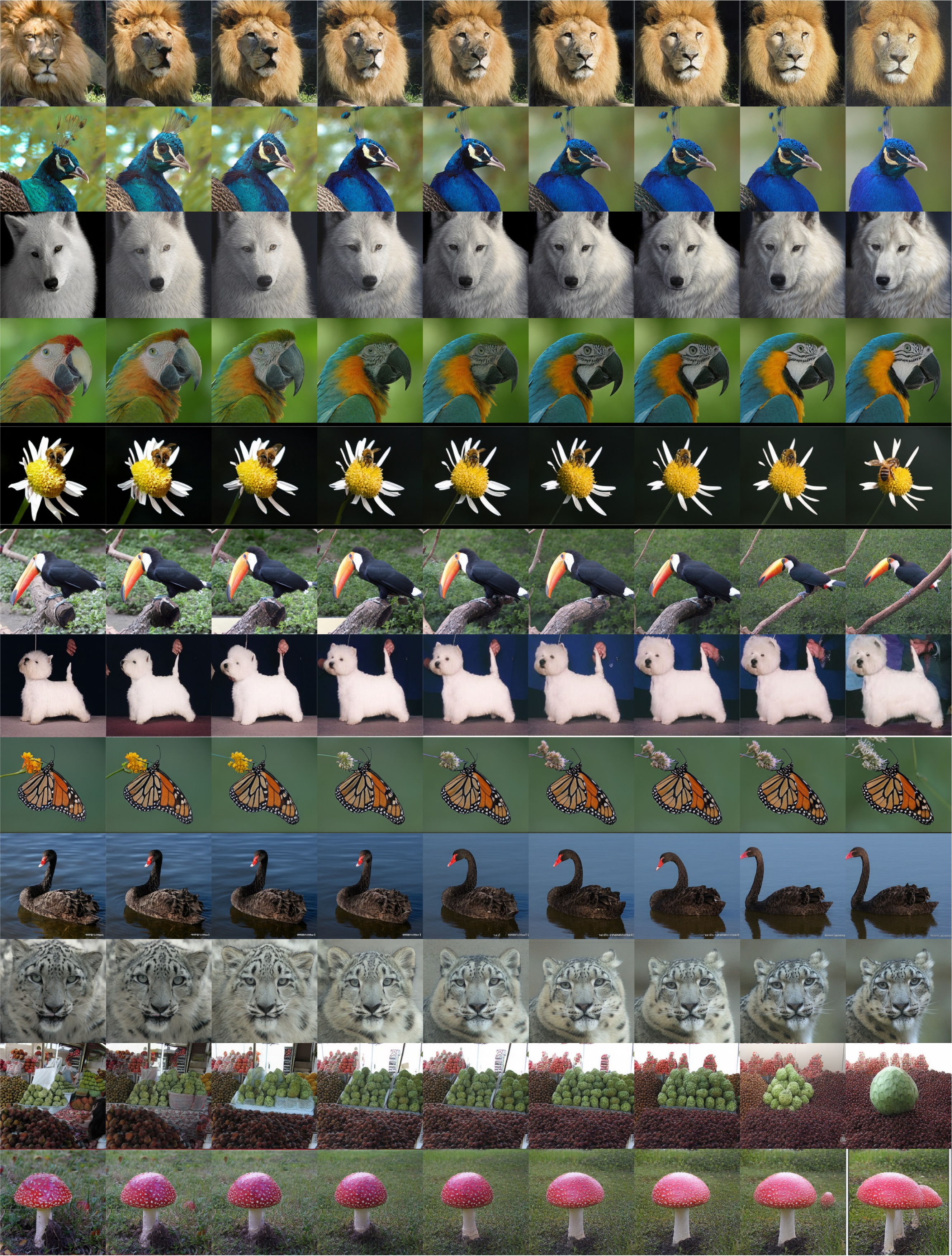}
    \caption{\textbf{Latent interpolation in the downstream DiT space.}
PAE supports smooth and semantically coherent transitions, indicating a learnable and stable latent space for diffusion.}
    \label{fig:dit_interp_vis}
\end{figure}

To assess the learnability and smoothness of the PAE latent manifold from the perspective of downstream diffusion, we visualize interpolation trajectories in the latent space of a trained DiT model in Figure~\ref{fig:dit_interp_vis}. By performing linear and spherical linear interpolation between two noise vectors, we observe that PAE maintains exceptional semantic coherence and image quality throughout the transition. The feature space of PAE is robust and continuous, supporting smooth semantic transitions that benefit both training stability and inference efficiency.

\subsection{Tokenizer Latent Interpolation}
\label{app:vae_interp}
\begin{figure}[t]
    \centering
    \includegraphics[width=\textwidth]{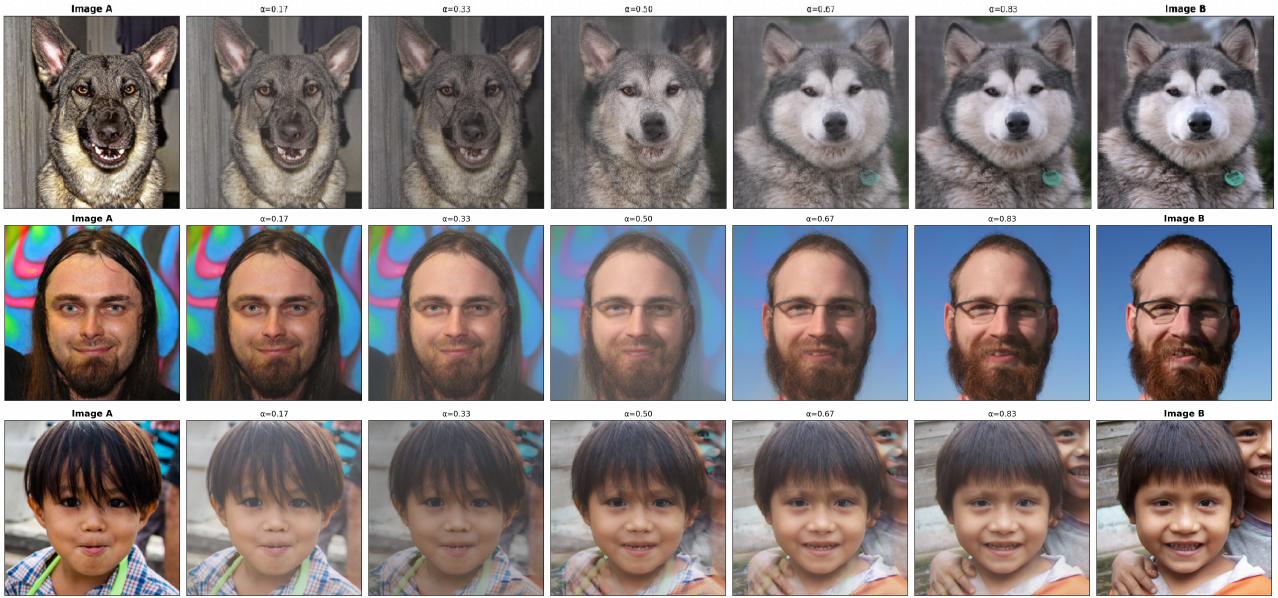}
    \caption{\textbf{Tokenizer latent interpolation.}
PAE exhibits smooth local transitions in latent space, consistent with improved local manifold continuity.}
    \label{fig:interp_vis1}
\end{figure}

\begin{figure}[t]
    \centering
    \includegraphics[width=\textwidth]{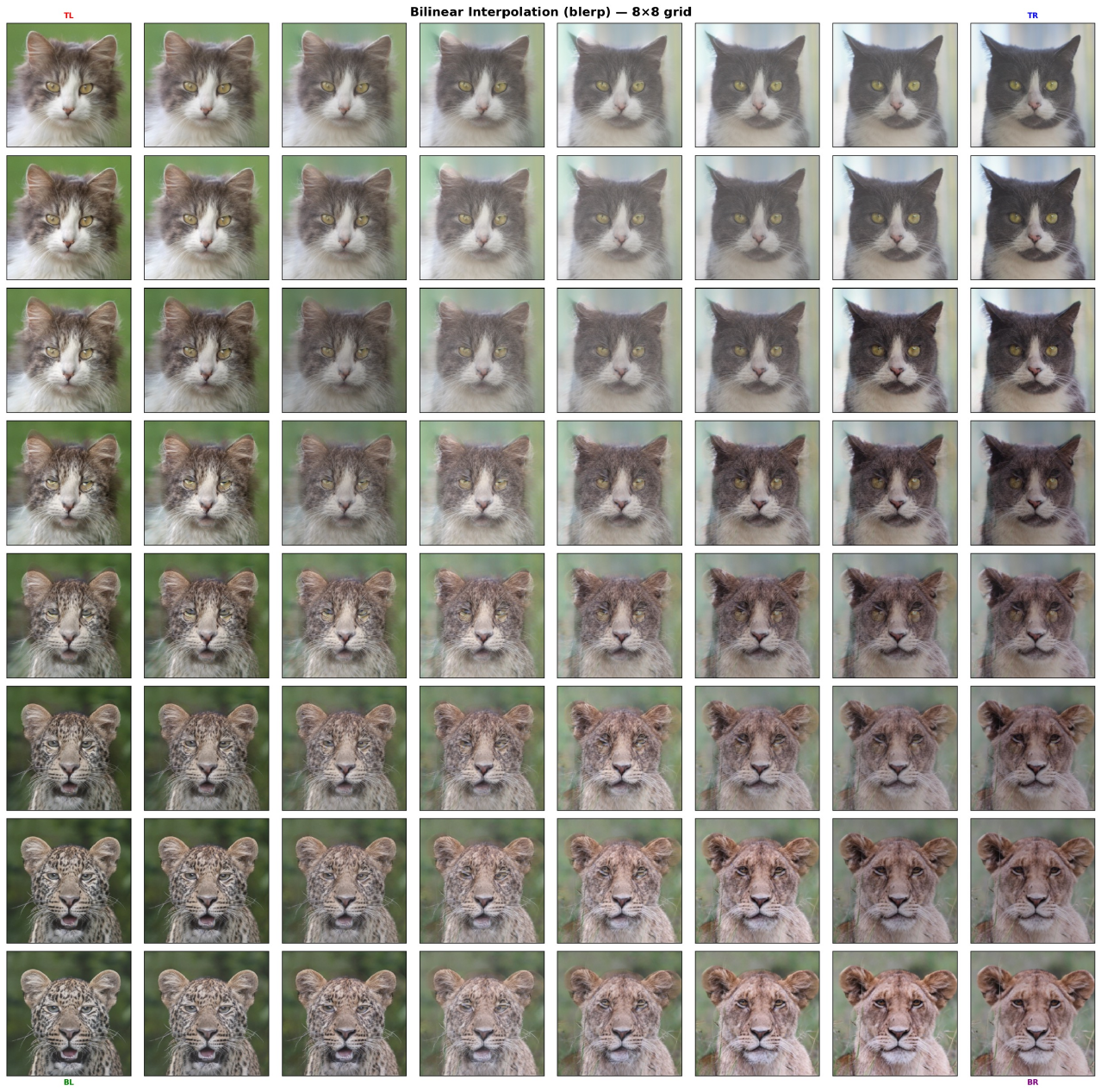}
    \caption{\textbf{Tokenizer latent interpolation.} Interpolation trajectories in PAE latent space illustrating local continuity.}
    \label{fig:interp_vis3}
\end{figure}

Figures~\ref{fig:interp_vis1} to \ref{fig:interp_vis3} further explore the local continuity of the tokenizer's latent space through direct interpolation between encoded image pairs. The smooth transitions across varied semantic categories (e.g., from one animal species to another) demonstrate that Manifold Continuity Regularization (MCR) effectively enforces a locally Lipschitz-continuous manifold. This local smoothness minimizes the LPC and ensures that small latent perturbations correspond to gradual perceptual changes in the pixel space.

\subsection{More Generation results}
\begin{figure}[t]
    \centering
    \includegraphics[width=\textwidth]{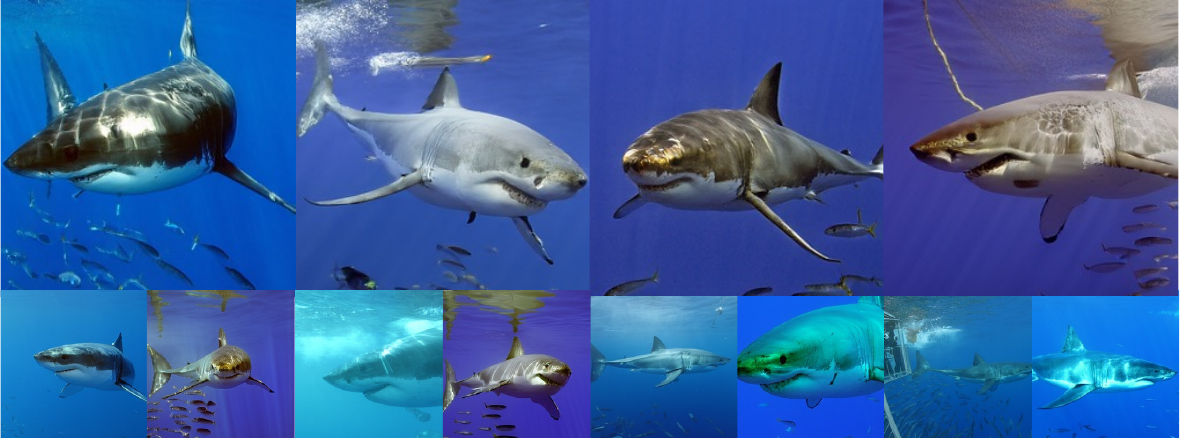}
    \caption{The visualization results of LightningDiT-XL/1 + PAE (DINOv2) use CFG with w = 3.3, and the class label is “Great white shark” (2).}
    \label{fig:class2}
\end{figure}

\begin{figure}[t]
    \centering
    \includegraphics[width=\textwidth]{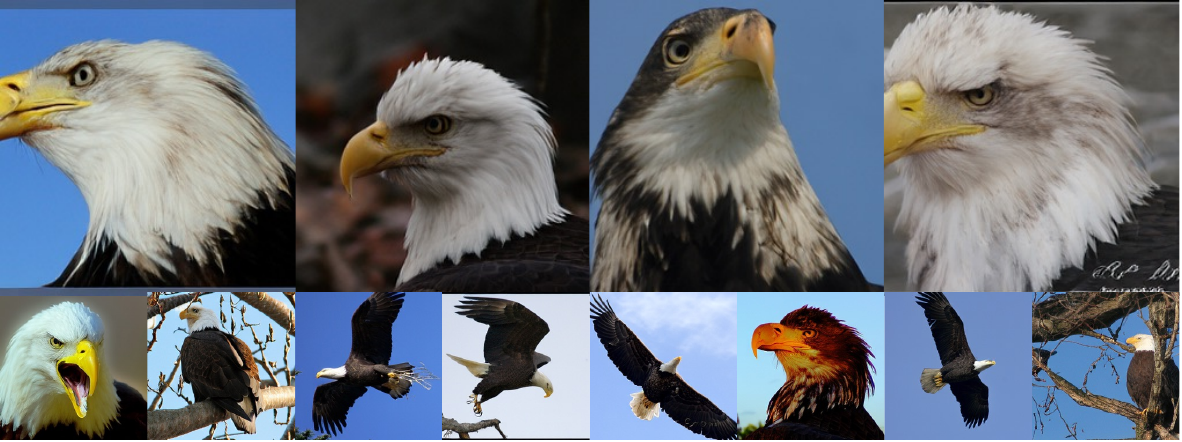}
    \caption{The visualization results of LightningDiT-XL/1 + PAE (DINOv2) use CFG with w = 3.3, and the class label is “Bald eagle” (22).}
    \label{fig:class22}
\end{figure}

\begin{figure}[t]
    \centering
    \includegraphics[width=\textwidth]{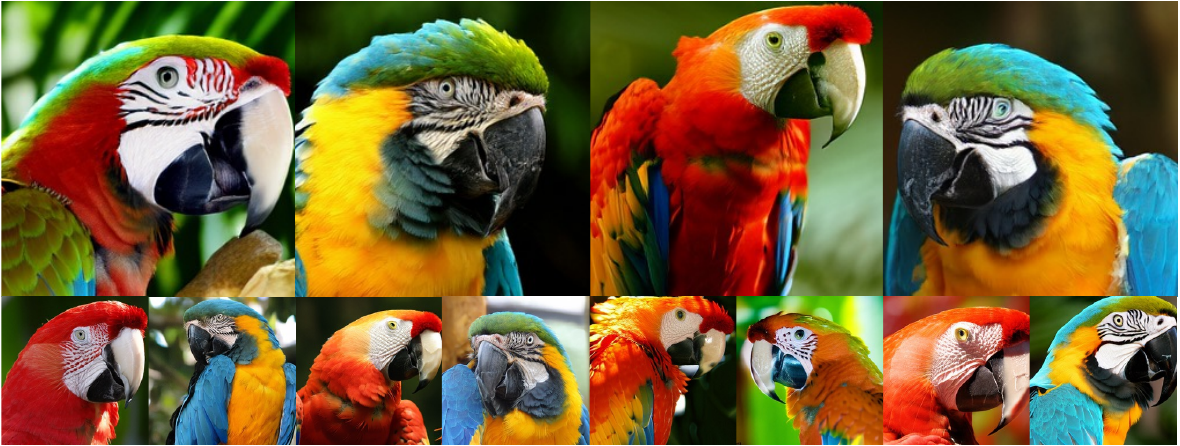}
    \caption{The visualization results of LightningDiT-XL/1 + PAE (DINOv2) use CFG with w = 3.3, and the class label is “Macaw” (88).}
    \label{fig:class88}
\end{figure}

\begin{figure}[t]
    \centering
    \includegraphics[width=\textwidth]{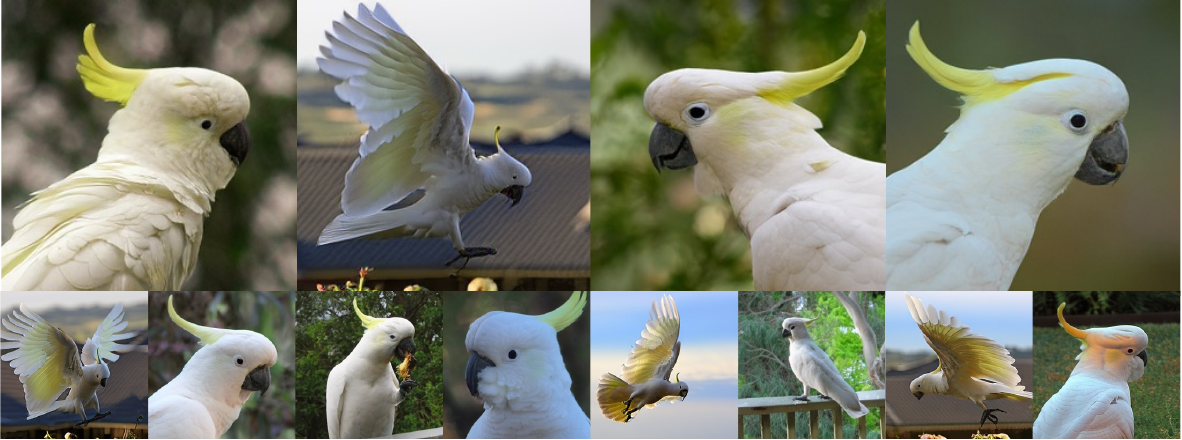}
    \caption{The visualization results of LightningDiT-XL/1 + PAE (DINOv2) use CFG with w = 3.3, and the class label is “Sulphur-crested cockatoo” (89).}
    \label{fig:class89}
\end{figure}

\begin{figure}[t]
    \centering
    \includegraphics[width=\textwidth]{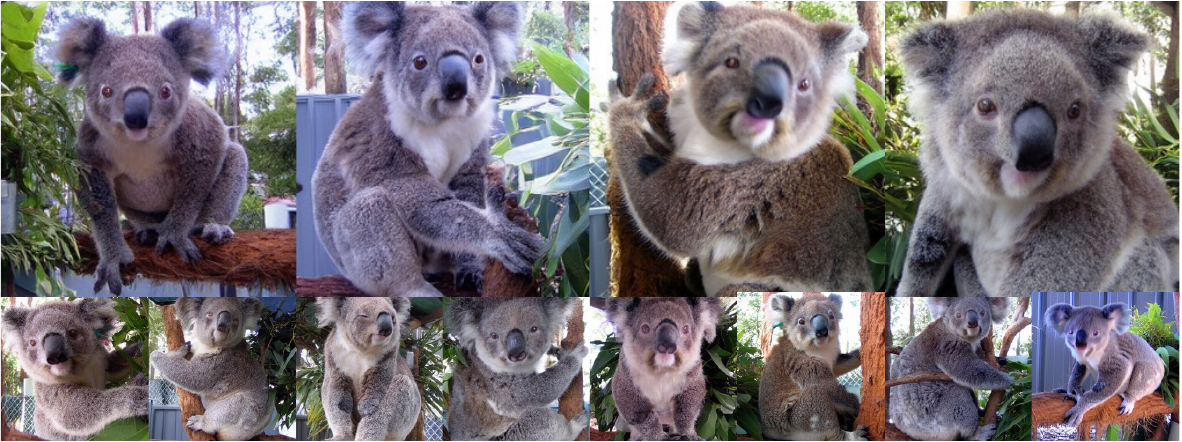}
    \caption{The visualization results of LightningDiT-XL/1 + PAE (DINOv2) use CFG with w = 3.3, and the class label is “Koala” (105).}
    \label{fig:class105}
\end{figure}

\begin{figure}[t]
    \centering
    \includegraphics[width=\textwidth]{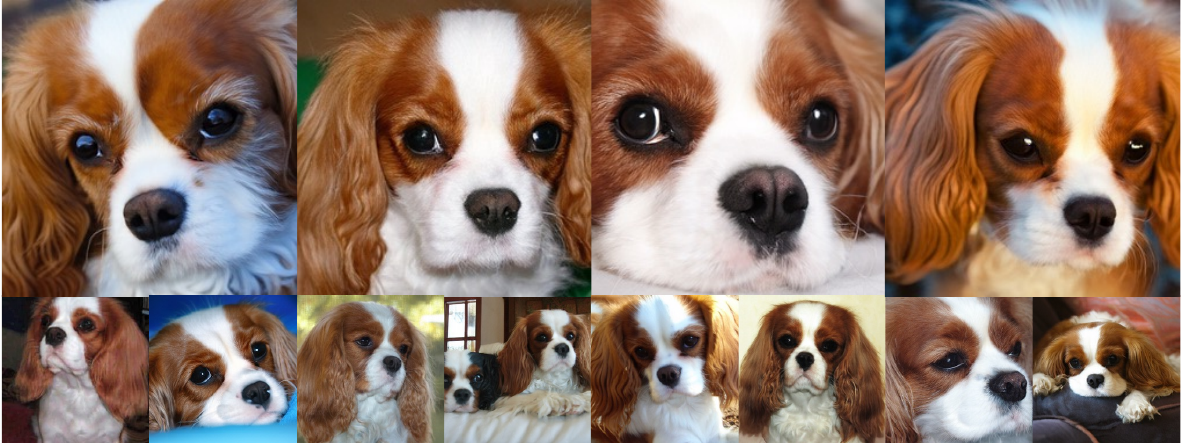}
    \caption{The visualization results of LightningDiT-XL/1 + PAE (DINOv2) use CFG with w = 3.3, and the class label is “Lesser panda” (156).}
    \label{fig:class156}
\end{figure}

\begin{figure}[t]
    \centering
    \includegraphics[width=\textwidth]{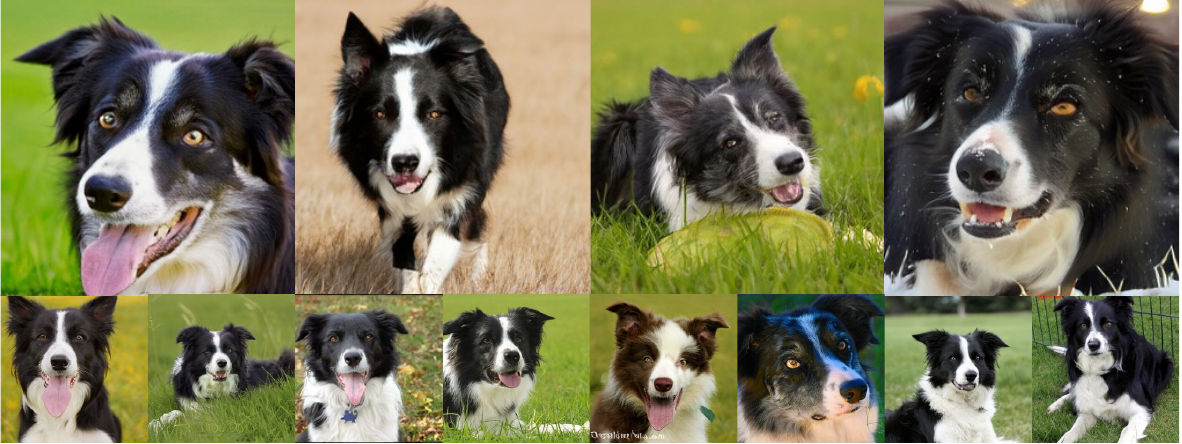}
    \caption{The visualization results of LightningDiT-XL/1 + PAE (DINOv2) use CFG with w = 3.3, and the class label is “Border collie” (232).}
    \label{fig:class232}
\end{figure}

\begin{figure}[t]
    \centering
    \includegraphics[width=\textwidth]{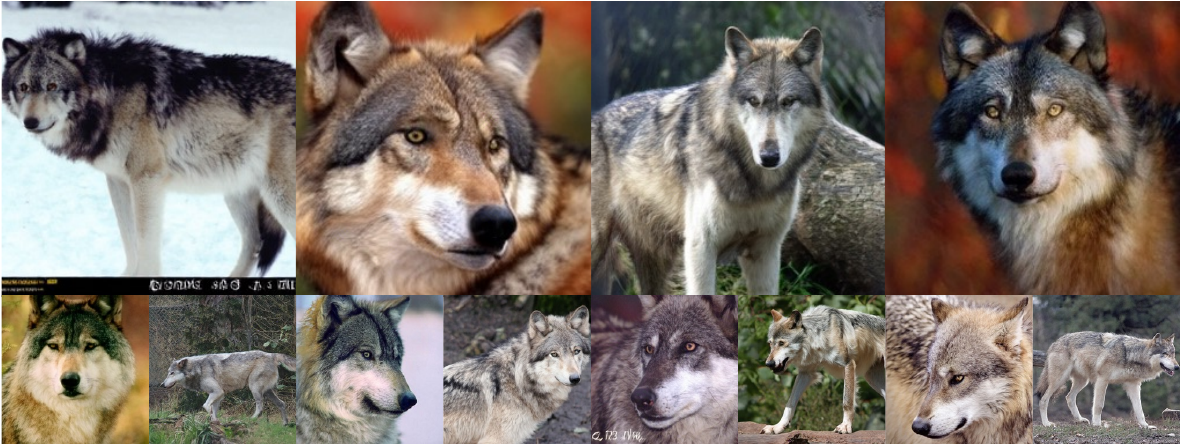}
    \caption{The visualization results of LightningDiT-XL/1 + PAE (DINOv2) use CFG with w = 3.3, and the class label is “Timber wolf” (269).}
    \label{fig:class269}
\end{figure}

\begin{figure}[t]
    \centering
    \includegraphics[width=\textwidth]{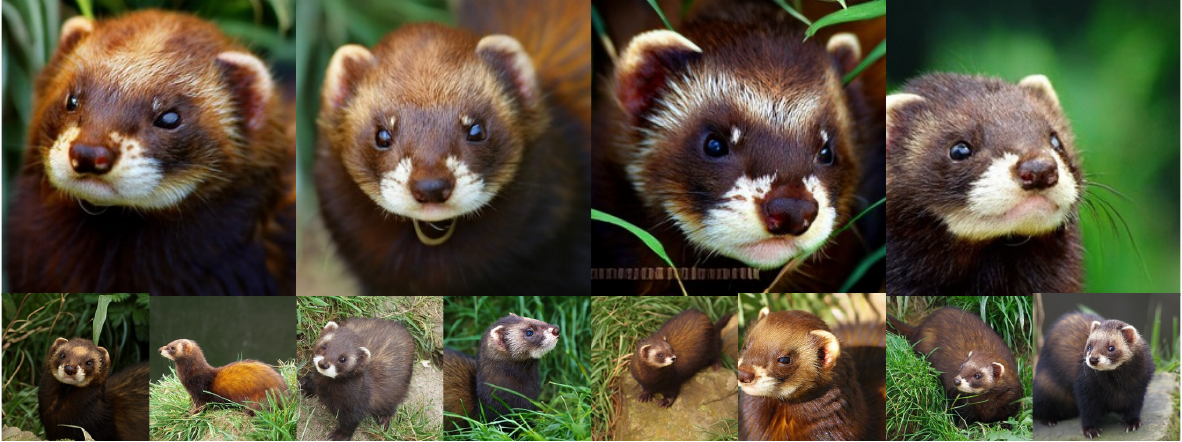}
    \caption{The visualization results of LightningDiT-XL/1 + PAE (DINOv2) use CFG with w = 3.3, and the class label is “Polecat” (358).}
    \label{fig:class358}
\end{figure}

\begin{figure}[t]
    \centering
    \includegraphics[width=\textwidth]{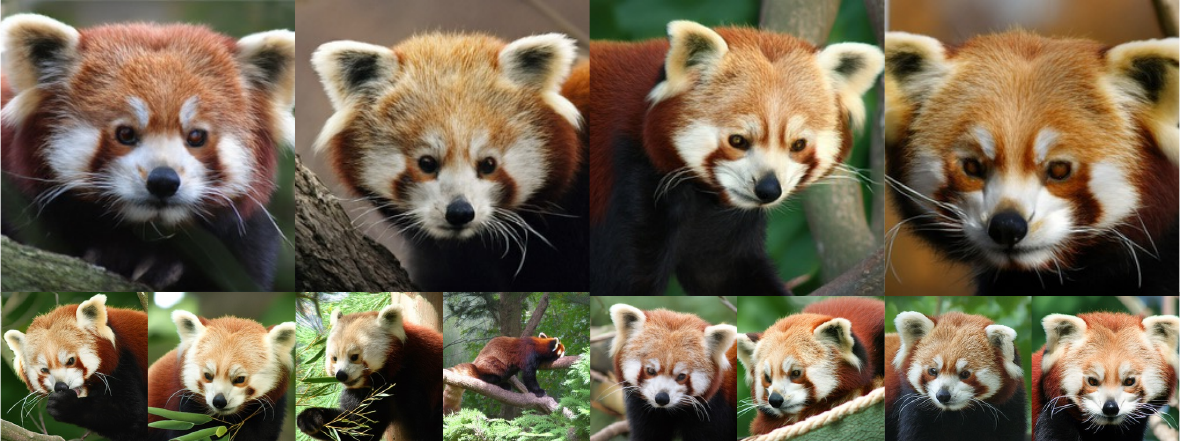}
    \caption{The visualization results of LightningDiT-XL/1 + PAE (DINOv2) use CFG with w = 3.3, and the class label is “Lesser panda” (387).}
    \label{fig:class387}
\end{figure}

\begin{figure}[t]
    \centering
    \includegraphics[width=\textwidth]{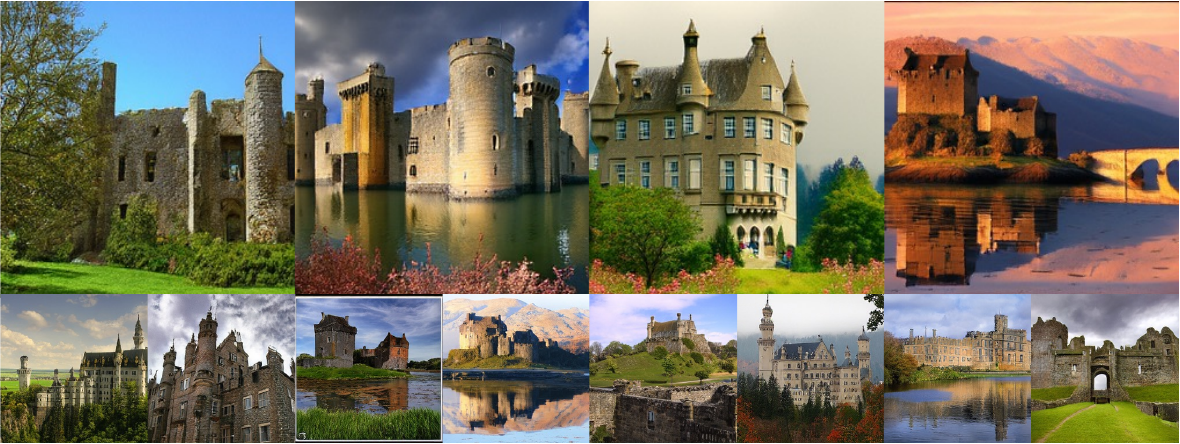}
    \caption{The visualization results of LightningDiT-XL/1 + PAE (DINOv2) use CFG with w = 3.3, and the class label is “Castle” (483).}
    \label{fig:class483}
\end{figure}

\begin{figure}[t]
    \centering
    \includegraphics[width=\textwidth]{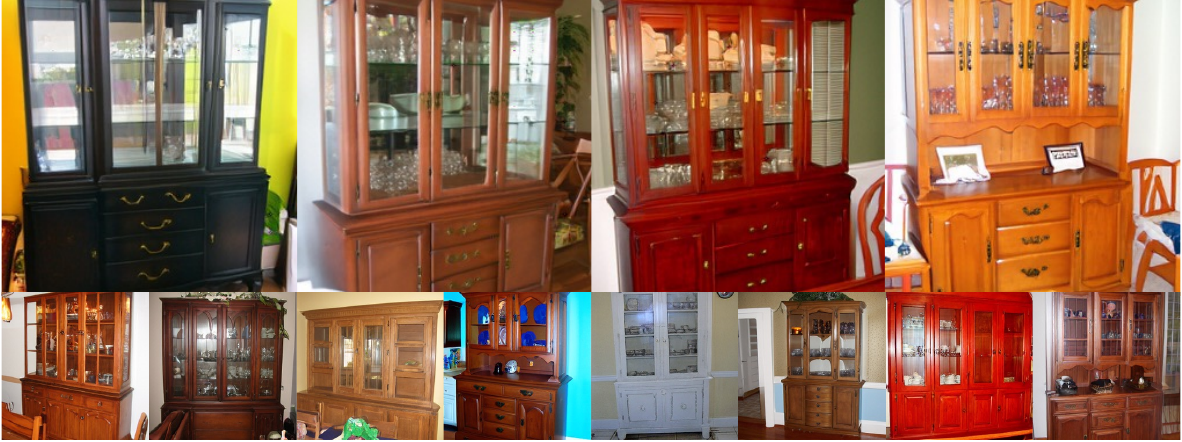}
    \caption{The visualization results of LightningDiT-XL/1 + PAE (DINOv2) use CFG with w = 3.3, and the class label is “China cabinet” (495).}
    \label{fig:class495}
\end{figure}

\begin{figure}[t]
    \centering
    \includegraphics[width=\textwidth]{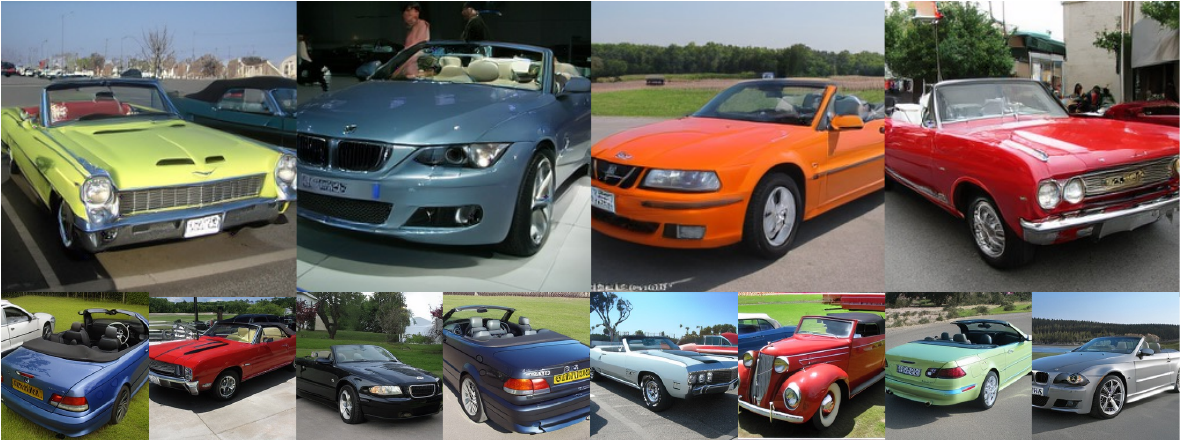}
    \caption{The visualization results of LightningDiT-XL/1 + PAE (DINOv2) use CFG with w = 3.3, and the class label is “Convertible” (511).}
    \label{fig:class511}
\end{figure}

\begin{figure}[t]
    \centering
    \includegraphics[width=\textwidth]{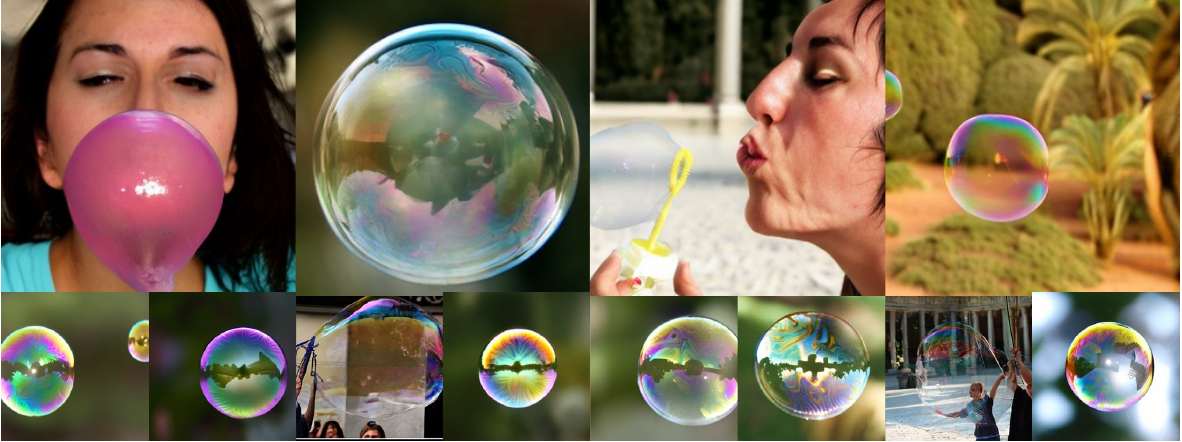}
    \caption{The visualization results of LightningDiT-XL/1 + PAE (DINOv2) use CFG with w = 3.3, and the class label is “Bubble” (971).}
    \label{fig:class971}
\end{figure}

\begin{figure}[t]
    \centering
    \includegraphics[width=\textwidth]{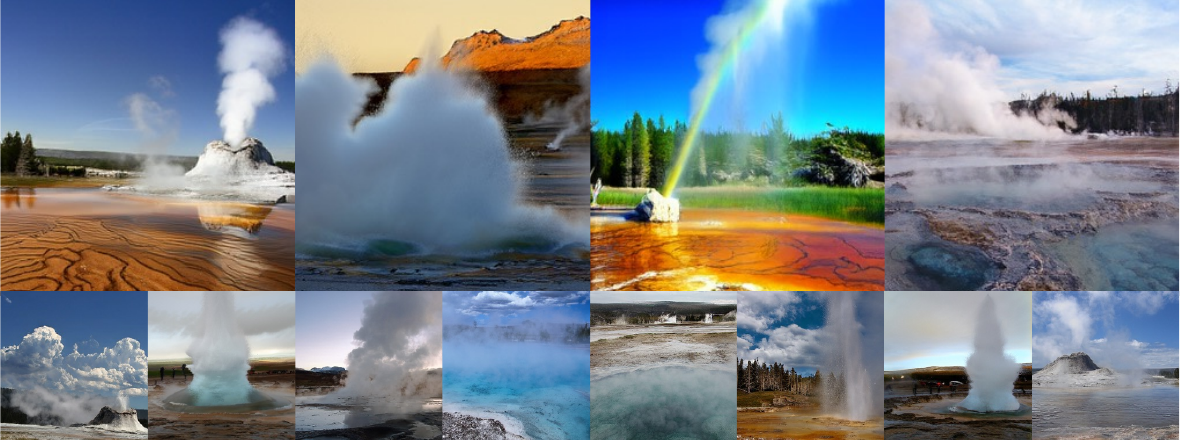}
    \caption{The visualization results of LightningDiT-XL/1 + PAE (DINOv2) use CFG with w = 3.3, and the class label is “Geyser” (974).}
    \label{fig:class974}
\end{figure}

We present more visualization results of PAE in Figs. \ref{fig:class2}–\ref{fig:class974} with CFG (w = 3.3).

\section{Limitations and Future Works}
Our current study is limited in several important aspects. First, all main experiments are conducted on ImageNet at $256{\times}256$ resolution, so the empirical conclusions are mainly validated in a single large-scale but relatively controlled setting. Although this setup is standard for tokenizer studies, it does not fully test whether the proposed method generalizes equally well to higher-resolution generation, more diverse visual domains, or downstream tasks beyond class-conditional image synthesis.

Second, our experiments focus on fixed-resolution latent diffusion. As a result, the current framework does not yet address settings with variable spatial scales, dynamic token allocation, or resolution-adaptive generation, where tokenizer design may interact more strongly with compression ratio and spatial capacity.

Third, while our results show that explicit prior alignment is effective for constructing a diffusion-friendly latent space, the current framework still relies on refined VFM-derived supervision and several carefully designed regularization terms. This leaves open the question of whether similar manifold properties could emerge more naturally from stronger tokenizer pretraining, larger-scale data, or more unified self-supervised objectives, without requiring explicit handcrafted alignment losses.

In future work, we plan to extend the study along these directions. In particular, we hope to evaluate the proposed perspective under larger-scale training, higher and dynamic resolutions, and broader generation settings, and to investigate whether stronger tokenizer pretraining can induce diffusion-friendly manifold organization more directly and robustly.

\section{Broader Impacts}
PAE provides a principled framework for rethinking tokenizer design in latent diffusion, showing how explicit latent-manifold organization can improve both generation quality and training efficiency beyond reconstruction-oriented objectives alone.